\documentclass[letterpaper]{article} 
\usepackage{_sty/aaai2027}  
\usepackage[hyphens]{url}  
\usepackage{graphicx} 
\usepackage{natbib}  
\usepackage{caption} 
\usepackage{amsmath}
\usepackage{amssymb}

\usepackage{algorithm}
\usepackage{algpseudocode}

\usepackage{booktabs}
\usepackage{graphicx}
\usepackage{booktabs}
\usepackage{multirow}
\usepackage{makecell}
\usepackage[table]{xcolor}

\usepackage{newfloat}
\usepackage{listings}
\DeclareCaptionStyle{ruled}{labelfont=normalfont,labelsep=colon,strut=off} 
\floatstyle{ruled}
\newfloat{listing}{tb}{lst}{}
\floatname{listing}{Listing}

\usepackage{booktabs}

\newbool{inccomment}
\booltrue{inccomment}  

\definecolor{mygray}{RGB}{230, 230, 230}

\newcommand{\XC}[1]{\ifbool{inccomment}{{\color{magenta}XC\@: #1}}{}}
\newcommand{\JY}[1]{\ifbool{inccomment}{{\color{blue}JY\@: #1}}{}}

\nocopyright

\title{Token Radius Attention for Efficient Video Generation}

\author{
    Jiayu Chen\textsuperscript{\rm 1}\equalcontrib,
    Zhikun Jiang\textsuperscript{\rm 2}\equalcontrib,
    Maoliang Li\textsuperscript{\rm 1},
    Jiayi Luo\textsuperscript{\rm 4,5},
    Jiawei Yang\textsuperscript{\rm 1},\\
    Zihao Zheng\textsuperscript{\rm 1},
    Hengyi Zhang\textsuperscript{\rm 1},
    Guojie Luo\textsuperscript{\rm 1,3},
    Xiang Chen\textsuperscript{\rm 1}\corresponding
}

\affiliations{
    \textsuperscript{\rm 1}School of Computer Science, Peking University\\
    \textsuperscript{\rm 2}School of Electronics Engineering and Computer Science, Peking University\\
    \textsuperscript{\rm 3}State Key Laboratory of Multimedia Information Processing, Peking University\\
    \textsuperscript{\rm 4}School of Computer Science and Engineering, Beihang University\\
    \textsuperscript{\rm 5}Zhongguancun Academy\\
    jiayu.chen.25@stu.pku.edu.cn,
    2400013188@stu.pku.edu.cn,
    xiang.chen@pku.edu.cn
}

\begin{document}

\maketitle

\begin{abstract}
Video Diffusion Transformers (VDiTs) enable high-fidelity generation but incur quadratic cost from dense 3D self-attention. Existing head- and block-level sparse methods share computation budgets across queries, overlooking token-specific attention demand. We observe that retained density varies across queries yet correlates log-linearly with attention entropy, while dominant interactions form query-centered neighborhoods with token-dependent radii. Based on these findings, we propose \textbf{Token Radius Attention (TRA)}, a training-free framework that maps query entropy to an analytic token budget and converts it into a temporally decayed radius without explicit key ranking. Fused entropy extraction, warm-up reuse, and block-sparse mask construction further reduce overhead. Across seven Wan2.1, Wan2.2, and HunyuanVideo T2V/I2V configurations, TRA retains only 9--19\% of attention interactions and achieves \(1.56\times\)--\(2.05\times\) speedup with competitive generation quality. Code is available at \url{https://github.com/IF-LAB-PKU/Token-Radius-Attention}.
\end{abstract}

\section{Introduction}
\label{sec:introduction}

Video Diffusion Transformers (VDiTs)~\cite{peebles2023scalable} have become a dominant architecture for high-fidelity video generation~\cite{wan2025wan,kong2024hunyuanvideo,yang2024cogvideox}, enabling applications such as audiovisual content creation and world modeling~\cite{seedance2026seedance,brooks2024video}. 
    Their success relies on dense 3D self-attention to capture spatial and temporal dependencies. 
    Given \(N\) video tokens, however, dense attention computes \(\mathcal{O}(N^2)\) pairwise interactions, causing its cost to grow rapidly with video duration and resolution. 
Attention has therefore become a major inference bottleneck for scaling and deployment.

Recent studies~\cite{zhang2026faster} reduce this cost by exploiting the intrinsic sparsity of attention and computing only potentially important interactions. 
    Based on the granularity of sparse decisions, existing methods can be broadly categorized into two groups. 
    \emph{Head-level} methods assign structured spatial or temporal patterns to individual attention heads~\cite{xi2025sparse,chen2026sparse,li2026radial}, whereas \emph{block-level} methods select important query--key block pairs~\cite{yang2026sparse,zhang2025spargeattention,luo2026attention}. 
Despite their effectiveness, both paradigms typically share a single computation allocation across all query tokens within a head or query block, implicitly assuming uniform attention demand within each group.

\begin{figure}[t]
    \centering
    \includegraphics[width=\columnwidth]{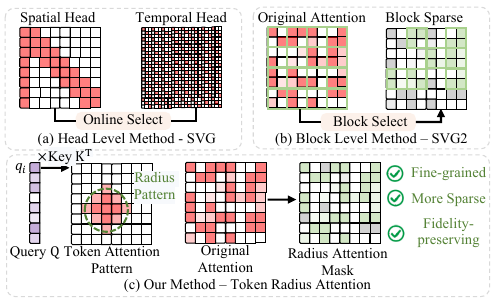}
    \caption{Comparison of sparse attention paradigms: TRA performs token-level radius masking for higher sparsity while preserving fidelity, unlike head-and block-level methods.}
    \label{fig:1}
\end{figure}

This shared-allocation assumption conflicts with the query token-specific computation of self-attention~\cite{vaswani2017attention}.
    Since softmax is normalized independently for each query, the number of keys required to preserve a target attention mass is inherently a \textbf{token-specific attention demand}. 
    Even within the same head or query block, concentrated queries require a small \textbf{token budget}, whereas diffuse queries require a much larger one.
    Consequently, shared head- or block-level allocation can waste computation on concentrated queries and miss critical keys for diffuse ones, degrading generation quality.

\emph{Token-level} sparse allocation can address this mismatch by aligning sparse decisions with the atomic unit of attention, but replacing shared allocation with token-specific decisions introduces an adaptivity--efficiency tension. 
    First, estimating and assigning an individual token budget to every query may itself incur substantial identification and decision overhead, calling for a low-cost model of token-specific attention demand. 
    Second, a budget specifies only how much computation to allocate, but not where to allocate it. 


To address this challenge, we analyze token-specific attention sparsity and its spatial support geometry. For each query, we define retained density as the minimum fraction of top-ranked keys required to preserve a target attention mass. Our analysis shows that retained density varies by orders of magnitude across queries within the same layer and head, while its logarithm grows approximately linearly with attention entropy, revealing token-specific attention sparsity (\textbf{Insight~I}). Accordingly, attention entropy provides a compact signal for assigning query-specific token budgets without requiring per-query key ranking. We further observe that high-attention keys form query-centered circular neighborhoods and normalized attention probability decays approximately exponentially with two-dimensional spatial distance, establishing the token radius attention pattern (\textbf{Insight~II}). Consequently, each budget can be converted into a structured spatial support through a token radius. With the temporal distance-decay rule~\cite{li2026radial}, these radii form regular spatiotemporal supports across video frames.

Based on these insights, we propose \textbf{Token Radius Attention (TRA)}, a training-free framework that realizes token-specific structured sparsity through an \emph{entropy-to-budget-to-radius} pipeline. Based on \textbf{Insight~I}, TRA uses an analytic entropy-based approximation to estimate a query-specific token budget, determining how much computation each query receives without per-query key ranking. During early dense warm-up, TRA computes attention entropy and reuses its budgets, radii, and masks throughout subsequent sparse steps. Based on \textbf{Insight~II}, TRA converts each predicted budget into a query-specific base radius and applies temporal distance decay to construct regular two-dimensional supports across video frames, determining where the computation is allocated. Finally, a tile-major token layout and fused CUDA kernel perform distance computation, radius comparison, block pruning, and token voting, converting logical token-radius masks into regular block masks for FlashInfer~\cite{ye2025flashinfer} execution. These components combine token-level adaptivity with efficient structured computation.

\begin{figure}[t]
    \centering
    \includegraphics[width=\columnwidth]{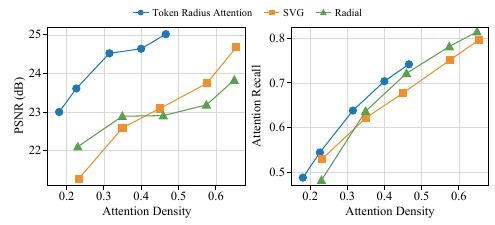}
    \caption{
    \textbf{PSNR and attention recall versus attention density.} Token Radius Attention achieves a favorable quality-efficiency trade-off compared with baselines.
    }
    \label{fig:2}
\end{figure}

Our contributions are threefold: (1) We uncover token-specific attention sparsity in VDiTs: retained density varies dramatically across queries yet follows a log-linear relationship with attention entropy, while interactions exhibit a query-centered token radius attention pattern. (2) We introduce \textbf{Token Radius Attention (TRA)}, a training-free entropy-to-budget-to-radius framework that transforms query-specific computation demand into regular spatiotemporal supports under temporal distance decay without explicit key ranking. (3) We co-design a radius-mask kernel and a fused entropy kernel, and comprehensively validate TRA across seven Wan2.1, Wan2.2, and HunyuanVideo variants spanning text-to-video and image-to-video generation, where it demonstrates a favorable quality--efficiency trade-off.

\section{Token Radius Attention}

\subsection{Preliminary and Challenge}
\label{sec:dense_attention}

Video Diffusion Transformers (VDiTs) represent a latent video using \(N=FHW\) tokens, where \(F\), \(H\), and \(W\) denote the temporal length and spatial grid dimensions, respectively. Each video token is associated with a spatiotemporal coordinate \((f,x,y)\). For a single self-attention head, let \(Q,K,V\in\mathbb{R}^{N\times d}\) denote the query, key, and value matrices, where \(d\) is the head dimension. Dense 3D self-attention computes all query--key interactions as
\begin{equation}
    \begin{aligned}
        s_{ij}&=\frac{q_i^{\top}k_j}{\sqrt{d}},
        &p_{ij}&=\frac{\exp(s_{ij})}{\sum_{n=1}^{N}\exp(s_{in})},
        o_i&=\sum_{j=1}^{N}p_{ij}v_j .
    \end{aligned}
    \label{eq:dense_attention}
\end{equation}
Although dense attention captures unrestricted spatial and temporal dependencies, it incurs \(\mathcal{O}(N^2)\) interactions. Sparse attention reduces this cost by retaining a subset of keys. \textbf{Challenge: }\textit{reconciling token-specific attention demand with structured sparse execution.} Existing head- or block-level methods share one retained density, radius schedule, or key-block set across multiple queries. However, this shared allocation conflicts with the row-wise nature of self-attention: because \(p_i=(p_{i1},\ldots,p_{iN})\) is normalized independently for each query \(i\), different queries need not require the same amount or extent of attention. Consequently, a large shared mask wastes computation on queries with concentrated support, whereas a small one may remove important interactions for queries with broader support. Addressing this challenge requires \textbf{token-specific structured sparsity}, with allocation adapting to each query while preserving a regular mask.

\subsection{Insight I: Token-Specific Attention Sparsity}
\label{sec:token_specific_sparsity}

We first determine \emph{how much} computation each query requires. For a target attention mass \(\tau\), let \(K_i^\tau\) be the minimum number of top-ranked keys whose cumulative probability reaches \(\tau\), and define the retained density as \(B_i^\tau=K_i^\tau/N\). Because obtaining \(K_i^\tau\) requires the complete attention row, it serves only as an oracle measure of query-specific attention demand. We therefore seek a compact statistic that exposes this demand without treating all queries uniformly.

To characterize the dispersion of query \(i\)'s attention distribution, we introduce its attention entropy:
\begin{equation}
    \mathcal{H}_i
    =
    -\sum_{j=1}^{N}p_{ij}\log p_{ij},
    \qquad
    0\leq\mathcal{H}_i\leq\log N.
    \label{eq:attention_entropy}
\end{equation}
A low entropy indicates that most attention mass is concentrated on a few keys, whereas a high entropy indicates that the mass is distributed over a broader support. Entropy therefore describes not merely whether attention is sparse, but how many interactions a particular query is likely to need.

\begin{figure}[t]
    \centering
    \includegraphics[width=\columnwidth]{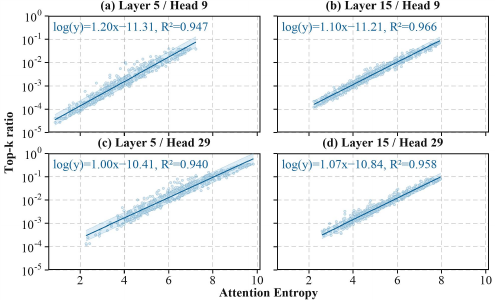}
    \caption{\textbf{Relationship between token-specific retained density and attention entropy on Wan2.2-A14B.} Their strong log-linear relationship (\(R^2=0.940\text{--}0.966\)) supports attention entropy as a query-specific budget signal.}
    \label{fig:3}
\end{figure}

Figure~\ref{fig:3} reveals two complementary properties. First, the retained density varies by orders of magnitude among tokens from the same layer and head, directly contradicting the shared-budget assumption. Second, \(\log B_i^\tau\) grows approximately linearly with \(\mathcal{H}_i\), with \(R^2=0.940\text{--}0.966\) across the four profiled layer--head groups. This log-linear trend is consistent with the effective-support interpretation of entropy: when attention spreads over more keys, its entropy increases logarithmically, while the number of keys needed to preserve the same mass grows exponentially. Thus, attention entropy serves as a compact signal for assigning query-specific token budgets without requiring a separately trained predictor.

\subsection{Insight II: Token Radius Attention Pattern}
\label{sec:token_radius_pattern}

A token budget specifies how many keys to retain, but it does not determine \emph{where} those keys should be placed. Selecting arbitrary top-ranked keys would require dense attention scores and an expensive per-query search. We therefore examine whether the dominant interactions of each video token follow a coordinate-defined pattern that can translate a scalar budget into structured support. For query \(i=(f_i,x_i,y_i)\) and key \(j=(f_j,x_j,y_j)\), we define
\begin{equation}
    d_s(i,j)=\sqrt{(x_i-x_j)^2+(y_i-y_j)^2},
    \qquad
    d_t(i,j)=|f_i-f_j|.
    \label{eq:spatiotemporal_distance}
\end{equation}

\begin{figure}[t]
    \centering
    \includegraphics[width=\columnwidth]{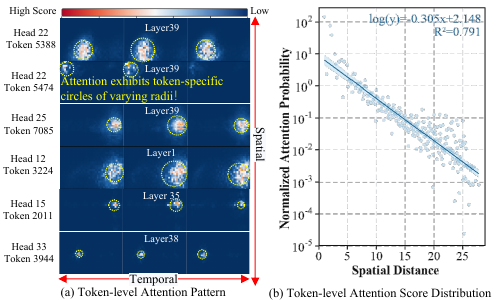}
    \caption{\textbf{Token-specific attention patterns on Wan2.2-A14B.} Dominant interactions form query-centered neighborhoods with token-dependent radii, while attention decays exponentially with two-dimensional spatial distance.}
    \label{fig:4}
\end{figure}

As shown in Figure~\ref{fig:4}(a), high-attention keys form approximately circular neighborhoods around the query's spatial position. Crucially, these neighborhoods are not identical: their radii vary substantially across queries, heads, and layers, mirroring the token-specific budgets identified above. Figure~\ref{fig:4}(b) explains why a radius is an effective support parameter. The normalized attention probability decays approximately exponentially with two-dimensional spatial distance (\(R^2=0.791\)); hence, enlarging a query-centered disk progressively incorporates lower-probability keys, whereas shrinking it removes distant interactions first. The radius therefore provides a monotonic, structured way to convert each predicted budget into a spatial support without ranking arbitrary key tokens.

The pattern also reflects the structure of video generation rather than sequence proximity. Within each frame, a video token attends most strongly to spatially nearby visual content; across frames, the query-centered layout provides a common coordinate system for aggregating locally aligned appearance and motion context. Because the useful extent varies with both the token and temporal offset, the spatial geometry can share an established temporal distance-decay rule~\cite{li2026radial}, while its base radius remains query specific.

\begin{figure*}[t]
    \centering
     \includegraphics[width=\textwidth]{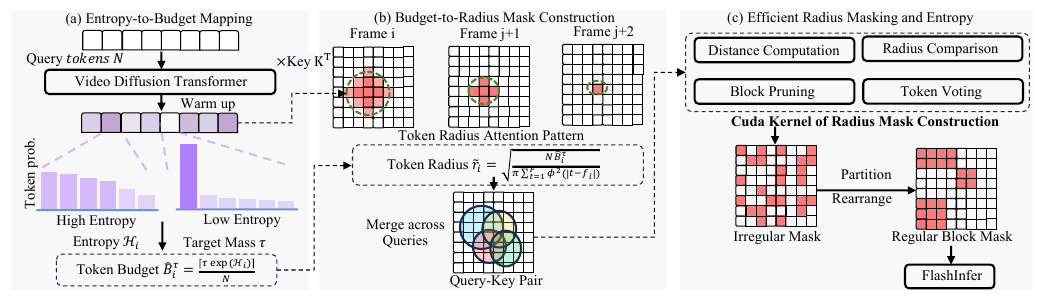}
    \caption{
    \textbf{Overview of Token Radius Attention (TRA).}
    (a) Query-wise entropy predicts a token budget.
    (b) The budget is converted into temporally decayed spatial radii.
    (c) A fused CUDA kernel constructs efficient block masks for execution.
}
    \label{fig:5}
\end{figure*}

\section{Efficient Video Generation with TRA}
\label{sec:method}

In this section, we introduce \textbf{Token Radius Attention (TRA)}, a training-free sparse attention framework that allocates computation at query-token granularity while retaining a regular execution structure for VDiTs. 
    As illustrated in Figure~\ref{fig:5}, TRA follows an \emph{entropy-to-budget-to-radius} pipeline. 
        First, entropy-guided token budgeting estimates a query-specific \emph{effective token budget} under a common attention-mass target (Section~\ref{sec:3_1}). 
        Second, a budget-preserving radius construction converts this scalar budget into a query-specific spatiotemporal mask without explicitly ranking individual keys (Section~\ref{sec:3_2}). Additionally, fused entropy extraction, cross-step reuse, and a customized block-sparse kernel amortize the decision overhead and translate token-level sparsity into practical execution (Section~\ref{sec:3_3}).

\subsection{Entropy-to-Budget Mapping}
\label{sec:3_1}

\paragraph{Token-Specific Budget Formulation.}
Following \textbf{Insight~I}, a low-entropy query concentrates its attention mass on a small number of key tokens, whereas a high-entropy query distributes its mass over a broader support. Consequently, different queries require different numbers of keys to preserve the same target attention mass. This heterogeneity motivates an entropy-based estimate of the effective token budget for each query.

Specifically, consider the attention distribution
\(\mathbf{p}_i=(p_{i1},\ldots,p_{iN})\in\mathbb{R}^{N}\)
of query \(i\), where \(N\) is the number of key tokens. Let \(\pi_i\) denote the permutation that sorts its probabilities in descending order. Given a target attention mass \(\tau\in(0,1)\), we define the oracle support size as
\begin{equation}
    K_i^{\tau}
    =
    \min
    \left\{
        k:
        \sum_{n=1}^{k}
        p_{i,\pi_i(n)}
        \geq \tau
    \right\}.
    \label{eq:oracle_token_budget}
\end{equation}
The corresponding \emph{effective token budget}, expressed as the fraction of retained keys, is \(B_i^{\tau}=K_i^{\tau}/N\). Although \(B_i^\tau\) directly characterizes query-specific attention demand, computing it requires the complete attention distribution and an explicit per-query top-\(k\) search. It therefore serves as an oracle diagnostic rather than an efficient inference-time policy.

\paragraph{Entropy-Equivalent Support Approximation.}
To motivate a tractable entropy-to-budget relationship, we introduce an idealized effective-support approximation. For query \(i\), consider an entropy-equivalent distribution \(\mathbf{p}'_i\) that assigns uniform probability to a set \(\mathcal{S}_i\) of \(m_i\) effective key tokens and zero probability elsewhere:
\begin{equation}
    p'_{ij}
    =
    \begin{cases}
        \frac{1}{m_i},
        & j\in\mathcal{S}_i,\\[2pt]
        0,
        & j\notin\mathcal{S}_i,
    \end{cases}
    \qquad
    |\mathcal{S}_i|=m_i.
    \label{eq:equivalent_uniform_distribution}
\end{equation}
Requiring \(\mathbf{p}'_i\) to have the same entropy \(\mathcal{H}_i\) as the original distribution gives
\begin{align}
    \mathcal{H}(\mathbf{p}'_i)
    &=
    -\sum_{j=1}^{N}
    p'_{ij}\log p'_{ij}
    \nonumber
    &=
    -\sum_{j\in\mathcal{S}_i}
    \frac{1}{m_i}
    \log\frac{1}{m_i}
    =
    \log m_i.
    \label{eq:uniform_support_entropy}
\end{align}
Hence, \(m_i=\exp(\mathcal{H}_i)\), which represents an entropy-equivalent effective support size rather than the exact Top-\(\tau\) support of the original distribution.

\paragraph{Entropy-Based Budget Approximation.}
From the entropy-equivalent distribution in Equation~\ref{eq:equivalent_uniform_distribution}, we define $m_i^{\mathrm{eff}}=\exp(\mathcal{H}_i)$
as the continuous effective support size of query \(i\), rather than the exact number of nonzero attention probabilities. Under the uniform-support approximation, the attention mass associated with \(k\) retained effective keys is approximated by
\begin{equation}
    \widetilde{M}_i(k)
    =
    \min\left\{
        \frac{k}{m_i^{\mathrm{eff}}},
        1
    \right\},
    \qquad
    k\in\{0,\ldots,N\}.
    \label{eq:uniform_cumulative_mass}
\end{equation}
Consequently, the estimated number of keys required to reach a target attention mass \(\tau\) is
\begin{equation}
    \widehat{K}_i^\tau
    =
    \min
    \left\{
        k:
        \widetilde{M}_i(k)\geq\tau
    \right\}
    =
    \left\lceil
        \tau\exp(\mathcal{H}_i)
    \right\rceil.
    \label{eq:entropy_predicted_support}
\end{equation}
The corresponding effective token budget is therefore
\begin{equation}
    \widehat{B}_i^\tau
    =
    \frac{\widehat{K}_i^\tau}{N}
    =
    \frac{
        \left\lceil
            \tau\exp(\mathcal{H}_i)
        \right\rceil
    }{N}
    \approx
    \frac{
        \tau\exp(\mathcal{H}_i)
    }{N}.
    \label{eq:entropy_budget_mapping}
\end{equation}

Equation~\ref{eq:entropy_budget_mapping} provides an analytic and monotonic approximation of the query-specific token budget under the effective-support assumption, assigning smaller budgets to low-entropy queries and larger budgets to high-entropy ones. It does not imply that entropy uniquely determines the exact Top-\(\tau\) support of the original attention distribution. Nevertheless, its exponential dependence is consistent with the empirical log-linear trend in Figure~\ref{fig:3}. Appendix further evaluates this approximation using measured attention distributions and provides the corresponding derivation. 

\subsection{Budget-to-Radius Mask Construction}
\label{sec:3_2}

\paragraph{Temporally Decayed Radius Model.}
The entropy-based budget approximation determines how many keys a query should retain, but not where those keys should be placed. Explicitly selecting the top-\(\widehat{K}_i^\tau\) keys would still require dense attention scores and a per-query ranking operation. Following \textbf{Insight~II}, TRA instead converts each budget into query-centered 2D disks, producing a coordinate-defined spatiotemporal support without arbitrary key selection.

Let query \(i\) be located at \((f_i,x_i,y_i)\), and let key \(j\) be located at \((f_j,x_j,y_j)\). We use the spatial and temporal distances defined in Equation~\ref{eq:spatiotemporal_distance}. Following the established temporal-locality prior~\cite{li2026radial}, we model the contraction of spatial support using
\begin{equation}
    \phi(\delta)
    =
    \exp(-\gamma\delta),
    \qquad
    \gamma\geq 0,
    \qquad
    \delta=|f_j-f_i|.
    \label{eq:temporal_radius_decay_final}
\end{equation}
Here, \(\gamma\) is a fixed decay rate shared by all queries; query-level adaptivity is introduced only through the base radius \(r_i\). The radius assigned to target frame \(t\) is
\(r_{i,t}=r_i\phi(|t-f_i|)\).
Thus, the predicted token budget controls the overall support scale, whereas temporal distance determines how this support is distributed across video frames.

\paragraph{Budget-Matched Radius Construction.}
The predicted density \(\widehat{B}_i^\tau\) corresponds to a target key count \(\widehat{K}_i^\tau=N\widehat{B}_i^\tau\). We determine a base radius whose frame-dependent disks contain approximately this number of video keys. Assuming latent-grid coordinates and temporarily ignoring discretization and spatial boundaries, the retained-key count is approximated by the sum of disk areas:
\begin{equation}
    C_i^{\mathrm{cont}}(r)
    =
    \sum_{t=1}^{F}
    \pi r^2\phi^2(|t-f_i|).
    \label{eq:continuous_radius_count_final}
\end{equation}
Equating \(C_i^{\mathrm{cont}}(r)\) with \(\widehat{K}_i^\tau\) yields the analytic initialization
\begin{equation}
    \widetilde{r}_i
    =
    \sqrt{
        \frac{
            \widehat{K}_i^\tau
        }{
            \pi
            \sum_{t=1}^{F}
            \phi^{2}(|t-f_i|)
        }
    }.
    \label{eq:analytic_radius_final}
\end{equation}
This formulation completes the entropy-to-budget-to-radius chain: entropy estimates a query-specific key count, while the temporal decay distributes spatial support across frames. However, \(\widetilde r_i\) is only a continuous approximation and does not account for finite grid boundaries or lattice effects.

To obtain a boundary-aware radius, we count the retained keys exactly for each candidate \(r\):
\begin{equation}
    C_i^{\mathrm{grid}}(r)
    =
    \sum_{t=1}^{F}
    \left|
        \left\{
            j:
            f_j=t,\;
            d_s(i,j)
            \leq
            r\phi(|t-f_i|)
        \right\}
    \right|.
    \label{eq:grid_radius_count_final}
\end{equation}
TRA selects the smallest candidate radius whose discrete support reaches the target key count:
\begin{equation}
    r_i
    =
    \min
    \left\{
        r\in\mathcal{R}:
        C_i^{\mathrm{grid}}(r)
        \geq
        \widehat{K}_i^\tau
    \right\},
    \label{eq:grid_radius_selection_final}
\end{equation}
where \(\mathcal{R}\) includes all candidate radii on the latent grid and a full-support radius. Because the count changes discretely, the selected support may slightly exceed \(\widehat{K}_i^\tau\); the construction therefore matches the predicted budget conservatively rather than exactly. For fixed grid dimensions and \(\gamma\), the boundary-aware counts depend only on the query coordinate, query frame, and candidate radius, allowing them to be stored in a lookup table.

\paragraph{Token-Radius Mask Definition.}
The resulting logical video-token mask is
\begin{equation}
    M_{ij}
    =
    \mathbb{I}
    \left[
        d_s(i,j)
        \leq
        r_i\phi(|f_j-f_i|)
    \right].
    \label{eq:token_radius_mask_final}
\end{equation}
The mask is query specific because \(r_i\) depends on the entropy-derived budget \(\widehat{K}_i^\tau\), yet structurally regular because its support is a union of nested two-dimensional disks. Conceptually, TRA applies the mask as
\begin{equation}
    \operatorname{TRA}(Q,K,V)
    =
    \operatorname{Softmax}
    \left(
        \frac{QK^\top}{\sqrt{d}}
        +
        \mathcal{B}(M)
    \right)V,
    \label{eq:tra_attention_final}
\end{equation}
where \(\mathcal{B}_{ij}(M)=0\) when \(M_{ij}=1\), and \(\mathcal{B}_{ij}(M)=-\infty\) otherwise. Equation~\ref{eq:tra_attention_final} is a mathematical definition; the sparse kernel evaluates only retained query--key blocks rather than materializing dense \(QK^\top\). For joint text--video attention, all text keys remain visible and \(M\) is applied only to the video-key submatrix; architectures with separate cross-attention retain their original text-attention path.

\subsection{Efficient Radius Masking and Entropy}
\label{sec:3_3}

\paragraph{Flash Radius Mask Kernel.}

Fixed-block sparse-attention kernels cannot represent Token-Radius masks because irregular regions do not align with fixed block boundaries. Under the original token order, tokens retained for one query may span multiple disjoint blocks, producing a fragmented sparse mask. TRA resolves this mismatch by partitioning each \(H\times W\) frame into \(b_h\times b_w\) tiles and rearranging video tokens in tile-major order. This layout makes neighboring tokens contiguous in the one-dimensional sequence, grouping retained interactions into fewer, denser blocks for efficient block-sparse computation. Before attention,\(Q\), \(K\), and \(V\) are consistently reordered; afterward, outputs are restored to the original token order.

Given this layout, TRA converts the query-specific token-level mask \(M\) into a hardware-compatible block-sparse mask. We fuse distance computation, radius comparison, block pruning, and token-level voting into one CUDA kernel. A three-dimensional grid covers all query blocks, key blocks, and attention heads, generating head-specific masks in one launch. For each query–key block pair, the kernel uses precomputed spatial bounding boxes to discard pairs that cannot contain valid token interactions. It then evaluates remaining candidates at token granularity and votes to determine whether each block is retained. Finally, all head masks are packed into one block-sparse attention invocation, avoiding per-head planning and dispatch overhead.

\paragraph{Fused Attention Entropy Kernel.}
Computing query-wise attention entropy at every denoising step would offset the benefit of sparse attention. TRA therefore extracts entropy only during selected early dense warm-up steps and reuses the resulting budgets, radii, and masks throughout the subsequent sparse steps. Specifically, letting \(s_{ij}=q_i^\top k_j/\sqrt{d}\), \(Z_i=\sum_j\exp(s_{ij})\), and \(p_{ij}=\exp(s_{ij})/Z_i\), the entropy can be rewritten as \(\mathcal{H}_i=\log Z_i-\sum_{j=1}^{N}p_{ij}s_{ij}\). This formulation allows entropy to be accumulated during online softmax without materializing the complete attention matrix. Following prior observations of cross-step entropy stability~\cite{chen2026ecovideo}, TRA directly reuses the entropy estimates from the final dense warm-up step to construct radius maps for subsequent sparse steps, reducing entropy estimation and mask-construction overhead.
\begin{table*}[!t]
  \centering
  \begingroup
    \fontsize{9}{10.8}\selectfont
    \setlength{\tabcolsep}{1.2pt}
    \renewcommand{\arraystretch}{0.95}
    \begin{tabular*}{\textwidth}{@{\extracolsep{\fill}}lccccccccccc@{}}
      \toprule
      Model/Method
      & \multicolumn{7}{c}{VBench (\%)}
      & \multicolumn{4}{c}{Efficiency} \\
      \cmidrule(lr){2-8}
      \cmidrule(lr){9-12}
      & Overall$\uparrow$
      & Subject$\uparrow$
      & Flicker$\uparrow$
      & Back.$\uparrow$
      & Aesthetic$\uparrow$
      & Motion$\uparrow$
      & Image$\uparrow$
      & Density$\downarrow$
      & FLOPs$\downarrow$
      & Latency$\downarrow$
      & Speedup$\uparrow$ \\
      \midrule

      \textit{Wan2.1-1.3B-T2V}
      & 87.67 & 97.56 & 99.55 & 97.93
      & 65.46 & 98.52 & 67.01
      & 1.00 & 105.34 & 417 & 1.000$\times$ \\

      SVG1
      & 84.44 & 94.27 & 98.75 & 91.16
      & 61.14 & 98.05 & 63.26
      & 0.30 & 48.41 & 266 & 1.568$\times$ \\

      SVG2
      & 85.30 & 95.88 & 98.75 & 96.47
      & 60.15 & 98.71 & 61.83
      & \underline{0.20}
      & \underline{40.22}
      & \underline{241}
      & \underline{1.730$\times$} \\

      Radial
      & \underline{86.30}
      & \underline{96.56}
      & \underline{99.05}
      & \underline{97.07}
      & \underline{62.76}
      & \underline{98.72}
      & \underline{63.67}
      & 0.39 & 55.31 & 257 & 1.623$\times$ \\

      \textbf{TRA}
      & \textbf{86.53}
      & \textbf{96.89}
      & \textbf{99.11}
      & \textbf{97.17}
      & \textbf{63.25}
      & \textbf{98.78}
      & \textbf{64.01}
      & \textbf{0.12}
      & \textbf{33.63}
      & \textbf{230}
      & \textbf{1.813$\times$} \\

      \midrule

      \textit{Wan2.1-14B-T2V}
      & 87.41 & 97.52 & 99.46 & 97.70
      & 61.27 & 99.08 & 69.43
      & 1.00 & 374.56 & 1982 & 1.000$\times$ \\

      SVG1
      & 86.45
      & 97.09
      & \underline{98.49}
      & 97.12
      & 58.95
      & 99.26
      & 67.80
      & \underline{0.30}
      & \underline{183.13}
      & \underline{1239}
      & \underline{1.600$\times$} \\

      SVG2
      & 86.52
      & 97.24
      & 98.29
      & 97.07
      & 59.06
      & 99.18
      & \underline{68.29}
      & 0.32 & 188.58 & 1261 & 1.572$\times$ \\

      Radial
      & \underline{86.83}
      & \textbf{97.55}
      & 98.16
      & \underline{97.49}
      & \textbf{60.83}
      & \textbf{99.28}
      & 67.66
      & 0.36 & 261.99 & 1297 & 1.528$\times$ \\

      \textbf{TRA}
      & \textbf{87.39}
      & \underline{97.43}
      & \textbf{99.10}
      & \textbf{97.61}
      & \underline{60.24}
      & \underline{99.27}
      & \textbf{71.86}
      & \textbf{0.15}
      & \textbf{141.93}
      & \textbf{1131}
      & \textbf{1.752$\times$} \\

      \midrule

      \textit{Wan2.2-14B-T2V}
      & 88.63 & 97.29 & 99.22 & 97.39
      & 67.22 & 98.94 & 71.75
      & 1.00 & 374.56 & 1608 & 1.000$\times$ \\

      SVG1
      & \underline{87.63}
      & 96.16
      & \underline{97.16}
      & \textbf{96.67}
      & \underline{65.13}
      & \underline{98.83}
      & \underline{71.84}
      & \underline{0.30}
      & \underline{183.02}
      & \underline{1049}
      & \underline{1.533$\times$} \\

      SVG2
      & 86.93
      & \textbf{96.22}
      & 96.83
      & 96.45
      & 62.17
      & 98.41
      & 71.51
      & 0.34 & 192.91 & 1061 & 1.516$\times$ \\

      Radial
      & 87.29
      & 95.78
      & 97.01
      & 96.22
      & 64.76
      & 98.48
      & 71.47
      & 0.36 & 261.99 & 1164 & 1.381$\times$ \\

      \textbf{TRA}
      & \textbf{87.91}
      & \underline{96.18}
      & \textbf{99.05}
      & \underline{96.46}
      & \textbf{65.27}
      & \textbf{99.55}
      & \textbf{71.96}
      & \textbf{0.15}
      & \textbf{142.99}
      & \textbf{1020}
      & \textbf{1.576$\times$} \\

      \midrule

      \textit{HunyuanVideo-13B-T2V}
      & 86.52 & 97.67 & 99.43 & 97.76
      & 57.28 & 99.45 & 67.56
      & 1.00 & 416.22 & 1783 & 1.000$\times$ \\

      SVG1
      & 85.55
      & 96.47
      & 98.43
      & 96.51
      & 55.56
      & 99.27
      & \underline{67.04}
      & 0.27
      & 193.55
      & \underline{897}
      & \underline{1.988$\times$} \\

      SVG2
      & 85.57
      & 96.78
      & 98.65
      & 96.49
      & 55.31
      & 99.31
      & 66.90
      & \underline{0.26}
      & 187.68
      & 909
      & 1.961$\times$ \\

      Radial
      & \underline{85.87}
      & \underline{97.04}
      & \underline{99.03}
      & \textbf{96.72}
      & \textbf{56.74}
      & \underline{99.35}
      & 66.33
      & 0.28
      & \underline{183.41}
      & 916
      & 1.947$\times$ \\

      \textbf{TRA}
      & \textbf{85.95}
      & \textbf{97.10}
      & \textbf{99.60}
      & \underline{96.58}
      & \underline{56.36}
      & \textbf{99.38}
      & \textbf{67.05}
      & \textbf{0.09}
      & \textbf{136.30}
      & \textbf{870}
      & \textbf{2.049$\times$} \\

      \midrule

      \textit{Wan2.1-14B-I2V}
      & 86.56 & 94.48 & 97.24 & 95.37
      & 61.86 & 98.91 & 71.52
      & 1.00 & 374.56 & 1658 & 1.000$\times$ \\

      SVG1
      & 85.97
      & 94.42
      & 96.68
      & 94.51
      & 61.02
      & 98.28
      & \underline{70.93}
      & 0.30
      & 183.13
      & 1047
      & 1.584$\times$ \\

      SVG2
      & \underline{86.36}
      & \textbf{95.72}
      & \underline{96.98}
      & 94.93
      & 61.45
      & 98.33
      & 70.76
      & \underline{0.29}
      & \underline{179.71}
      & \underline{998}
      & \underline{1.661$\times$} \\

      Radial
      & 86.30
      & 94.52
      & 96.95
      & \underline{95.36}
      & \textbf{61.62}
      & \underline{98.44}
      & 70.91
      & 0.36
      & 261.99
      & 1046
      & 1.585$\times$ \\

      \textbf{TRA}
      & \textbf{86.41}
      & \underline{95.55}
      & \textbf{97.15}
      & \textbf{95.42}
      & \underline{61.59}
      & \textbf{98.74}
      & \textbf{71.09}
      & \textbf{0.19}
      & \textbf{151.96}
      & \textbf{971}
      & \textbf{1.708$\times$} \\

      \midrule

      \textit{Wan2.2-14B-I2V}
      & 88.60 & 97.55 & 97.54 & 97.24
      & 66.23 & 98.67 & 74.36
      & 1.00 & 374.56 & 1605 & 1.000$\times$ \\

      SVG1
      & \underline{87.40}
      & \textbf{97.17}
      & 97.01
      & \underline{97.08}
      & 62.05
      & 98.23
      & \textbf{72.88}
      & 0.30
      & 183.03
      & \underline{1034}
      & \underline{1.552$\times$} \\

      SVG2
      & 87.16
      & 96.79
      & 97.04
      & 96.65
      & 62.36
      & \textbf{98.84}
      & 71.28
      & \underline{0.28}
      & \underline{176.48}
      & 1057
      & 1.518$\times$ \\

      Radial
      & 87.10
      & 95.91
      & \textbf{97.48}
      & 96.49
      & \underline{63.73}
      & 97.80
      & 71.19
      & 0.36
      & 261.99
      & 1157
      & 1.387$\times$ \\

      \textbf{TRA}
      & \textbf{87.87}
      & \underline{97.10}
      & \underline{97.36}
      & \textbf{97.11}
      & \textbf{64.50}
      & \underline{98.70}
      & \underline{72.45}
      & \textbf{0.15}
      & \textbf{143.00}
      & \textbf{1028}
      & \textbf{1.561$\times$} \\

      \midrule

      \textit{HunyuanVideo-13B-I2V}
      & 87.30 & 96.55 & 98.91 & 96.54
      & 62.06 & 99.45 & 70.30
      & 1.00 & 416.22 & 1761 & 1.000$\times$ \\

      SVG1
      & 85.31
      & \textbf{96.51}
      & 98.15
      & 95.47
      & 55.57
      & 99.11
      & 67.04
      & \underline{0.27}
      & 193.55
      & \underline{887}
      & \underline{1.985$\times$} \\

      SVG2
      & 86.14
      & 95.61
      & \underline{98.28}
      & \underline{95.68}
      & 59.75
      & 99.06
      & 68.45
      & \underline{0.27}
      & 192.07
      & 889
      & 1.981$\times$ \\

      Radial
      & \underline{86.48}
      & 95.30
      & 97.58
      & 95.62
      & \textbf{61.66}
      & \underline{99.21}
      & \textbf{69.52}
      & 0.28
      & \underline{183.41}
      & 912
      & 1.931$\times$ \\

      \textbf{TRA}
      & \textbf{86.63}
      & \underline{96.17}
      & \textbf{98.41}
      & \textbf{96.01}
      & \underline{60.73}
      & \textbf{99.38}
      & \underline{69.41}
      & \textbf{0.09}
      & \textbf{136.30}
      & \textbf{874}
      & \textbf{2.015$\times$} \\

      \bottomrule
    \end{tabular*}
  \endgroup
  \normalfont\normalsize
  \caption{Quality and efficiency comparisons of TRA and training-free sparse-attention baselines on T2V and I2V tasks. VBench scores are reported in percentage, FLOPs in P, and latency in seconds. Speedup is measured against the corresponding dense model. Best and second-best sparse results are bolded and underlined, respectively.}
  \label{tab:main_results}
\end{table*}
\section{Experiments}

\subsection{Experimental settings}
\paragraph{Models and tasks.}
We evaluate TRA on seven T2V and I2V configurations based on Wan2.1, Wan2.2, and HunyuanVideo, covering model sizes from 1.3B to 14B. Unless otherwise specified, videos are generated at \(720 \times 1280\) resolution.
\paragraph{Datasets.}
For T2V evaluation, we use the prompt-enhanced Penguin benchmark provided by the VBench team~\cite{huang2024vbench}. For I2V evaluation, we use the 16:9 prompt--image pairs from VBench++~\cite{zheng2025vbench}.
\paragraph{Evaluation metrics and baselines.}
Generation quality is evaluated using VBench Overall and six dimensions: Subject Consistency, Temporal Flickering, Background Consistency, Aesthetic Quality, Motion Smoothness, and Imaging Quality. We report attention density, FLOPs, end-to-end latency, and speedup for efficiency evaluation. We measure fidelity to Dense Attention using PSNR, SSIM~\cite{wang2004image}, and LPIPS~\cite{zhang2018unreasonable}. We compare TRA with three training-free sparse attention methods: SVG~\cite{xi2025sparse}, SVG2~\cite{yang2026sparse}, and Radial ~\cite{li2026radial}.

\begin{figure*}[!t]
    \vspace{-6mm}
    \centering
    \includegraphics[width=\textwidth]{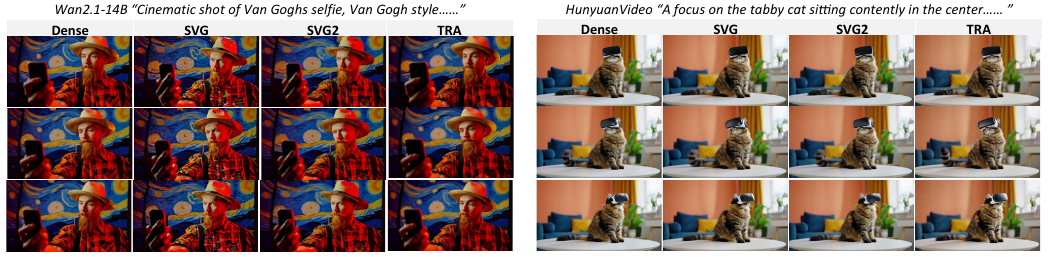}
    \caption{Qualitative comparison of various methods on Wan2.1 and HunyuanVideo.}
    \vspace{-2mm}
    \label{fig:6}

\end{figure*}

\paragraph{Implementation details.}
All experiments are conducted on an NVIDIA H200 GPU. At the default resolution, Wan2.1-1.3B, Wan2.1-14B, and Wan2.2-14B generate 81 frames, whereas HunyuanVideo-13B generates 129 frames. For TRA, SVG, and SVG2, the first transformer layer remains dense, and dense attention is applied during the first 25\% of the denoising process as warm-up. Method-specific sparse attention is used in the remaining layers and denoising steps. TRA uses a temporal decay factor of $0.6$ for all Wan models and $0.95$ for HunyuanVideo. SVG adopts its calibrated target block density, while SVG2 uses $K_q=300$ and $K_k=1000$; both follow the same layer-wise and denoising-step warm-up protocol as TRA. For Radial Attention, due to constraints imposed by its acceleration strategy, we use a resolution of $768 \times 1280$ and generate 69, 77, and 117 frames for Wan2.1, Wan2.2, and HunyuanVideo, respectively.

\begin{table}[t]
    \centering
    \fontsize{9}{9.5}\selectfont
    \setlength{\tabcolsep}{12pt}
    \renewcommand{\arraystretch}{0.90}
    \begin{tabular}{@{}lccc@{}}
        \toprule
        \textbf{Method}
        & \textbf{PSNR} $\uparrow$
        & \textbf{SSIM} $\uparrow$
        & \textbf{LPIPS} $\downarrow$ \\
        \midrule
        Wan2.1-14B
        & -- & -- & -- \\
        SVG
        & 21.2930 & 0.7980 & 0.2100 \\
        Radial
        & 20.4520 & 0.7020 & 0.2700 \\
        \midrule
        TRA
        & \textbf{23.3772} & \textbf{0.8275} & \textbf{0.1742} \\
        w/o Budget Mapping
        & 22.2070 & 0.5648 & 0.4403 \\
        w/o Radius Masking
        & 13.1320 & 0.2724 & 0.7011 \\
        \bottomrule
    \end{tabular}
    \caption{Ablation study on the key components of TRA.}
    \label{tab:2}
\end{table}
\begin{table}[t]
    \centering
    \fontsize{9}{9.5}\selectfont
    \setlength{\tabcolsep}{8pt}
    \renewcommand{\arraystretch}{0.90}
    \begin{tabular}{@{}lccc@{}}
        \toprule
        \textbf{Operation / Time (s)}
        & \textbf{Per Layer}
        & \textbf{Total}
        & \textbf{Speedup} \\
        \midrule
        Naive Token Radius Mask
        & 15.78 & 615.30 & 1.00$\times$ \\
        Flash Radius Mask
        & 1.52 & 59.46 & 10.35$\times$ \\
        Naive Attention Entropy
        & 4.77 & 190.91 & 1.00$\times$ \\
        Fused Attention Entropy
        & 0.39 & 15.47 & 12.34$\times$ \\
        \bottomrule
    \end{tabular}
    \caption{Efficiency evaluation for customized kernels.}
    \vspace{-6mm}
    \label{tab:3}
\end{table}

\subsection{Main Results}
\paragraph{Quantitative Comparison on VBench.}
We evaluate TRA on seven T2V and I2V configurations in Table~\ref{tab:main_results}. TRA achieves the best VBench Overall score among sparse methods in all seven configurations. Notably, it nearly matches dense attention on Wan2.1-14B T2V while achieving a $1.75\times$ speedup, showing a favorable quality--efficiency trade-off.

\paragraph{Efficiency Analysis.}
TRA achieves the highest speedup across all configurations, ranging from $1.56\times$ to $2.05\times$, while TRA retains only 9\%–19\% of attention interactions. On Hunyuan-13B, it reaches $2.05\times$ and $2.02\times$ speedups for T2V and I2V, respectively. TRA also consistently reduces FLOPs compared with SVG and Radial Attention, confirming the effectiveness of token-level adaptive sparsity.

\paragraph{Qualitative Visualizations of Different Methods.}
Figure~\ref{fig:6} compares TRA with dense attention and existing sparse methods. TRA better preserves subject identity, structural integrity, fine-grained details, and temporal consistency, while reducing artifacts such as distortion, flickering, and motion discontinuity. Its outputs remain visually close to dense attention despite substantial acceleration.

\subsection{Ablation Studies}

\textbf{Impact of Key Components.} As shown in Table~\ref{tab:2}, replacing entropy-guided allocation with a uniform budget degrades generation quality. Replacing the budget-preserving radius with a 1D radial distance causes a more severe degradation, including a PSNR drop of over 10 dB. These results confirm that the two components jointly enable TRA to allocate computation adaptively while preserving informative tokens.

\textbf{Computational Cost Analysis.}As shown in Table~\ref{tab:3}, Flash Radius Mask and Fused Attention Entropy achieve $10.35\times$ and $12.34\times$ speedups over their naive implementations, respectively. This demonstrates that the customized kernels substantially reduce the auxiliary cost of TRA, enabling its efficient practical deployment.

\textit{\textbf{Additional ablations are provided in the Appendix.}}
\section{Related Work}

\paragraph{Efficient Video Generation.}
Existing methods accelerate video diffusion by caching computation across denoising steps~\cite{liu2025timestep,liu2025reusing,ma2026magcache}, reducing sampling steps through distillation~\cite{salimans2022progressive,yin2024one}, or lowering per-evaluation cost through quantization~\cite{shang2023post,he2023ptqd,feng2025s}, and parallelism. These techniques optimize repeated evaluations or numerical cost, whereas TRA reduces the quadratic query--key interactions within executed attention layers and is therefore complementary.

\paragraph{Sparse Attention for Video Generation.}
Training-free sparse attention methods for video DiTs can be categorized by their allocation granularity. Head-level methods, including SVG~\cite{xi2025sparse}, Radial Attention~\cite{li2026radial}, and Sparse-vDiT~\cite{chen2026sparse}, assign structured spatial or temporal patterns to entire heads. Block-level methods, such as SpargeAttn~\cite{zhang2025spargeattention}, SVG2~\cite{yang2026sparse}, and SVOO~\cite{luo2026attention}, select important query--key block pairs. Both paradigms typically share one sparse allocation across queries, overlooking query-specific attention demand. TRA instead maps each query's entropy to its own token budget and spatiotemporal radius, combining token-level allocation with regular block-sparse execution.

\section{Conclusion}

We presented Token Radius Attention (TRA), a training-free framework that maps attention entropy to query-specific token budgets and temporally decayed radii. Across seven 720P T2V and I2V settings, TRA retains only \(9\%\)--\(19\%\) attention density, achieves \(1.56\times\)--\(2.05\times\) speedup, and delivers the competitive VBench score in seven settings.


\bibliography{_ref/aaai2027}

@String(CVPR= {IEEE Conf. Comput. Vis. Pattern Recog.})

@String(ICCV= {Int. Conf. Comput. Vis.})

@String(ECCV= {Eur. Conf. Comput. Vis.})

@String(NIPS= {Adv. Neural Inform. Process. Syst.})

@String(AAAI = {AAAI})

@String(CVPR  = {CVPR})

@String(ICCV  = {ICCV})

@String(ECCV  = {ECCV})

@String(NIPS  = {NeurIPS})

@String(ICML  = {ICML})

@article{vaswani2017attention,
  title={Attention is all you need},
  author={Vaswani, A},
  journal=NIPS,
  year={2017}
}

@article{salimans2022progressive,
  title={Progressive distillation for fast sampling of diffusion models},
  author={Salimans, Tim and Ho, Jonathan},
  journal={arXiv preprint arXiv:2202.00512},
  year={2022}
}

@article{xi2025sparse,
  title={Sparse VideoGen: Accelerating Video Diffusion Transformers with Spatial-Temporal Sparsity},
  author={Xi, Haocheng and Yang, Shuo and Zhao, Yilong and Xu, Chenfeng and Li, Muyang and Li, Xiuyu and Lin, Yujun and Cai, Han and Zhang, Jintao and Li, Dacheng and others},
  journal={arXiv preprint arXiv:2502.01776},
  year={2025}
}

@article{kong2024hunyuanvideo,
    title = {{HunyuanVideo}: a systematic framework for large video generative models},
    volume = {abs/2412.03603},
    url = {https://arxiv.org/abs/2412.03603},
    journal = {CoRR},
    author = {Kong, Weijie and Tian, Qi and Zhang, Zijian and Min, Rox and Dai, Zuozhuo and Zhou, Jin and {others}},
    year = {2024},
}

@inproceedings{peebles2023scalable,
  title={Scalable diffusion models with transformers},
  author={Peebles, William and Xie, Saining},
  booktitle={Proceedings of the IEEE/CVF international conference on computer vision},
  pages={4195--4205},
  year={2023}
}

@article{wan2025wan,
  title={Wan: Open and advanced large-scale video generative models},
  author={Wan, Team and Wang, Ang and Ai, Baole and Wen, Bin and Mao, Chaojie and Xie, Chen-Wei and Chen, Di and Yu, Feiwu and Zhao, Haiming and Yang, Jianxiao and others},
  journal={arXiv preprint arXiv:2503.20314},
  year={2025}
}

@article{yang2024cogvideox,
  title={Cogvideox: Text-to-video diffusion models with an expert transformer},
  author={Yang, Zhuoyi and Teng, Jiayan and Zheng, Wendi and Ding, Ming and Huang, Shiyu and Xu, Jiazheng and Yang, Yuanming and Hong, Wenyi and Zhang, Xiaohan and Feng, Guanyu and others},
  journal={arXiv preprint arXiv:2408.06072},
  year={2024}
}

@article{brooks2024video,
  title={Video generation models as world simulators},
  author={Brooks, Tim and Peebles, Bill and Holmes, Connor and DePue, Will and Guo, Yufei and Jing, Leo and Schnurr, David and Taylor, Joe and Luhman, Troy and Luhman, Eric and others},
  journal={OpenAI Blog},
  volume={1},
  number={8},
  pages={1},
  year={2024}
}

@article{seedance2026seedance,
  title={Seedance 2.0: Advancing video generation for world complexity},
  author={Seedance, Team and Chen, De and Chen, Liyang and Chen, Xin and Chen, Ying and Chen, Zhuo and Chen, Zhuowei and Cheng, Feng and Cheng, Tianheng and Cheng, Yufeng and others},
  journal={arXiv preprint arXiv:2604.14148},
  year={2026}
}

@article{zhang2026faster,
  title={Faster video diffusion with trainable sparse attention},
  author={Zhang, Peiyuan and Chen, Yongqi and Huang, Haofeng and Lin, Will and Liu, Zhengzhong and Stoica, Ion and Xing, Eric and Zhang, Hao},
  journal=NIPS,
  volume={38},
  pages={152509--152534},
  year={2026}
}

@article{zhang2025spargeattention,
  title={Spargeattention: Accurate and training-free sparse attention accelerating any model inference},
  author={Zhang, Jintao and Xiang, Chendong and Huang, Haofeng and Wei, Jia and Xi, Haocheng and Zhu, Jun and Chen, Jianfei},
  journal=ICML,
  year={2025}
}

@article{yang2026sparse,
  title={Sparse videogen2: Accelerate video generation with sparse attention via semantic-aware permutation},
  author={Yang, Shuo and Xi, Haocheng and Zhao, Yilong and Li, Muyang and Zhang, Jintao and Cai, Han and Lin, Yujun and Li, Xiuyu and Xu, Chenfeng and Peng, Kelly and others},
  journal=NIPS,
  volume={38},
  pages={96965--96991},
  year={2026}
}

@article{li2026radial,
  title={Radial Attention: $O(nlogn)$ Sparse Attention with Energy Decay for Long Video Generation},
  author={Li, Xingyang and Li, Muyang and Cai, Tianle and Xi, Haocheng and Yang, Shuo and Lin, Yujun and Zhang, Lvmin and Yang, Songlin and Hu, Jinbo and Peng, Kelly and others},
  journal=NIPS,
  volume={38},
  pages={16822--16852},
  year={2026}
}

@article{luo2026attention,
  title={Attention Sparsity is Input-Stable: Training-Free Sparse Attention for Video Generation via Offline Sparsity Profiling and Online QK Co-Clustering},
  author={Luo, Jiayi and Chen, Jiayu and Wang, Jiankun and Wang, Cong and Zhu, Hanxin and Sun, Qingyun and Gao, Chen and Chen, Zhibo and Li, Jianxin},
  journal=ICML,
  year={2026}
}

@inproceedings{chen2026sparse,
  title={Sparse-vdit: Unleashing the power of sparse attention to accelerate video diffusion transformers},
  author={Chen, Pengtao and Zeng, Xianfang and Zhao, Maosen and Shen, Mingzhu and Cheng, Wei and Yu, Gang and Chen, Tao},
  booktitle=AAAI,
  volume={40},
  number={4},
  pages={2957--2965},
  year={2026}
}

@article{chen2026ecovideo,
  title={EcoVideo: Entropy-Orchestrated Video Generation Paradigm in Cloud-Edge Dynamics},
  author={Chen, Jiayu and Zhang, Hengyi and Li, Maoliang and Li, Minyu and Zheng, Zihao and Liu, Xuanzhe and Luo, Guojie and Chen, Xiang},
  journal=ECCV,
  year={2026}
}

@inproceedings{huang2024vbench,
  title={Vbench: Comprehensive benchmark suite for video generative models},
  author={Huang, Ziqi and He, Yinan and Yu, Jiashuo and Zhang, Fan and Si, Chenyang and Jiang, Yuming and Zhang, Yuanhan and Wu, Tianxing and Jin, Qingyang and Chanpaisit, Nattapol and others},
  booktitle=CVPR,
  pages={21807--21818},
  year={2024}
}

@article{zheng2025vbench,
  title={Vbench-2.0: Advancing video generation benchmark suite for intrinsic faithfulness},
  author={Zheng, Dian and Huang, Ziqi and Liu, Hongbo and Zou, Kai and He, Yinan and Zhang, Fan and Gu, Lulu and Zhang, Yuanhan and He, Jingwen and Zheng, Wei-Shi and others},
  journal={arXiv preprint arXiv:2503.21755},
  year={2025}
}

@inproceedings{zhang2018unreasonable,
  title={The unreasonable effectiveness of deep features as a perceptual metric},
  author={Zhang, Richard and Isola, Phillip and Efros, Alexei A and Shechtman, Eli and Wang, Oliver},
  booktitle=CVPR,
  pages={586--595},
  year={2018}
}

@article{wang2004image,
  title={Image quality assessment: from error visibility to structural similarity},
  author={Wang, Zhou and Bovik, Alan C and Sheikh, Hamid R and Simoncelli, Eero P},
  journal={IEEE transactions on image processing},
  volume={13},
  number={4},
  pages={600--612},
  year={2004},
  publisher={IEEE}
}

@inproceedings{liu2025timestep,
  title={Timestep Embedding Tells: It's Time to Cache for Video Diffusion Model},
  author={Liu, Feng and Zhang, Shiwei and Wang, Xiaofeng and Wei, Yujie and Qiu, Haonan and Zhao, Yuzhong and Zhang, Yingya and Ye, Qixiang and Wan, Fang},
  booktitle=CVPR,
  pages={7353--7363},
  year={2025}
}

@article{ma2026magcache,
  title={Magcache: Fast video generation with magnitude-aware cache},
  author={Ma, Zehong and Wei, Longhui and Wang, Feng and Zhang, Shiliang and Tian, Qi},
  journal=NIPS,
  volume={38},
  pages={34348--34380},
  year={2026}
}

@inproceedings{liu2025reusing,
  title={From reusing to forecasting: Accelerating diffusion models with taylorseers},
  author={Liu, Jiacheng and Zou, Chang and Lyu, Yuanhuiyi and Chen, Junjie and Zhang, Linfeng},
  booktitle=ICCV,
  pages={15853--15863},
  year={2025}
}

@inproceedings{yin2024one,
  title={One-step diffusion with distribution matching distillation},
  author={Yin, Tianwei and Gharbi, Micha{\"e}l and Zhang, Richard and Shechtman, Eli and Durand, Fredo and Freeman, William T and Park, Taesung},
  booktitle={Proceedings of the IEEE/CVF conference on computer vision and pattern recognition},
  pages={6613--6623},
  year={2024}
}

@article{feng2025s,
  title={$S^2$Q-VDiT: Accurate Quantized Video Diffusion Transformer with Salient Data and Sparse Token Distillation},
  author={Feng, Weilun and Qin, Haotong and Yang, Chuanguang and Li, Xiangqi and Yang, Han and Li, Yuqi and An, Zhulin and Huang, Libo and Magno, Michele and Xu, Yongjun},
  journal={arXiv preprint arXiv:2508.04016},
  year={2025}
}

@article{he2023ptqd,
  title={Ptqd: Accurate post-training quantization for diffusion models},
  author={He, Yefei and Liu, Luping and Liu, Jing and Wu, Weijia and Zhou, Hong and Zhuang, Bohan},
  journal=NIPS,
  volume={36},
  pages={13237--13249},
  year={2023}
}

@inproceedings{shang2023post,
  title={Post-training quantization on diffusion models},
  author={Shang, Yuzhang and Yuan, Zhihang and Xie, Bin and Wu, Bingzhe and Yan, Yan},
  booktitle=CVPR,
  pages={1972--1981},
  year={2023}
}

@article{ye2025flashinfer,
    title = {FlashInfer: Efficient and Customizable Attention Engine for LLM Inference Serving},
    author = {
      Ye, Zihao and
      Chen, Lequn and
      Lai, Ruihang and
      Lin, Wuwei and
      Zhang, Yineng and
      Wang, Stephanie and
      Chen, Tianqi and
      Kasikci, Baris and
      Grover, Vinod and
      Krishnamurthy, Arvind and
      Ceze, Luis
    },
    journal = {arXiv preprint arXiv:2501.01005},
    year = {2025}
}


\clearpage
\appendix
\raggedbottom
\setcounter{topnumber}{2}
\setcounter{dbltopnumber}{2}
\renewcommand{\topfraction}{0.95}
\renewcommand{\dbltopfraction}{0.95}
\renewcommand{\textfraction}{0.05}
\renewcommand{\floatpagefraction}{0.85}
\renewcommand{\dblfloatpagefraction}{0.85}
\setlength{\textfloatsep}{8pt plus 2pt minus 2pt}
\setlength{\dbltextfloatsep}{8pt plus 2pt minus 2pt}
\setlength{\abovedisplayskip}{6pt plus 2pt minus 2pt}
\setlength{\belowdisplayskip}{6pt plus 2pt minus 2pt}

\section{Supplementary Analyses}
\label{app:supplementary_analyses}

\subsection{Cross-Model Validation on HunyuanVideo}
\label{app:hunyuan_validation}

\paragraph{Analysis setup.}
TRA is motivated by two empirical properties of video self-attention:
query-specific retained density is strongly correlated with attention
entropy, and dominant query--key interactions follow a query-centered Token
Radius Attention pattern. To verify that these properties are not specific
to the Wan architecture, we repeat the analyses on
HunyuanVideo-13B~\cite{kong2024hunyuanvideo}.

\begin{figure*}[!t]
    \centering
    \begin{minipage}[t]{0.49\textwidth}
        \centering
        \includegraphics[width=\linewidth]{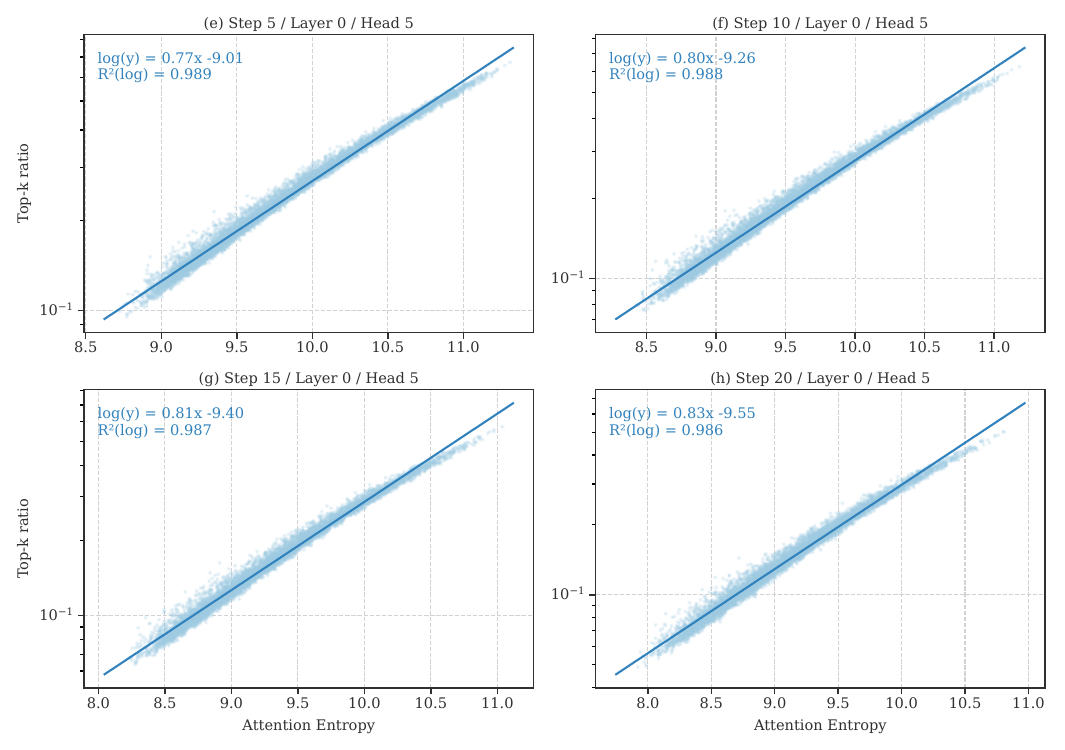}
    \end{minipage}
    \hfill
    \begin{minipage}[t]{0.49\textwidth}
        \centering
        \includegraphics[width=\linewidth]{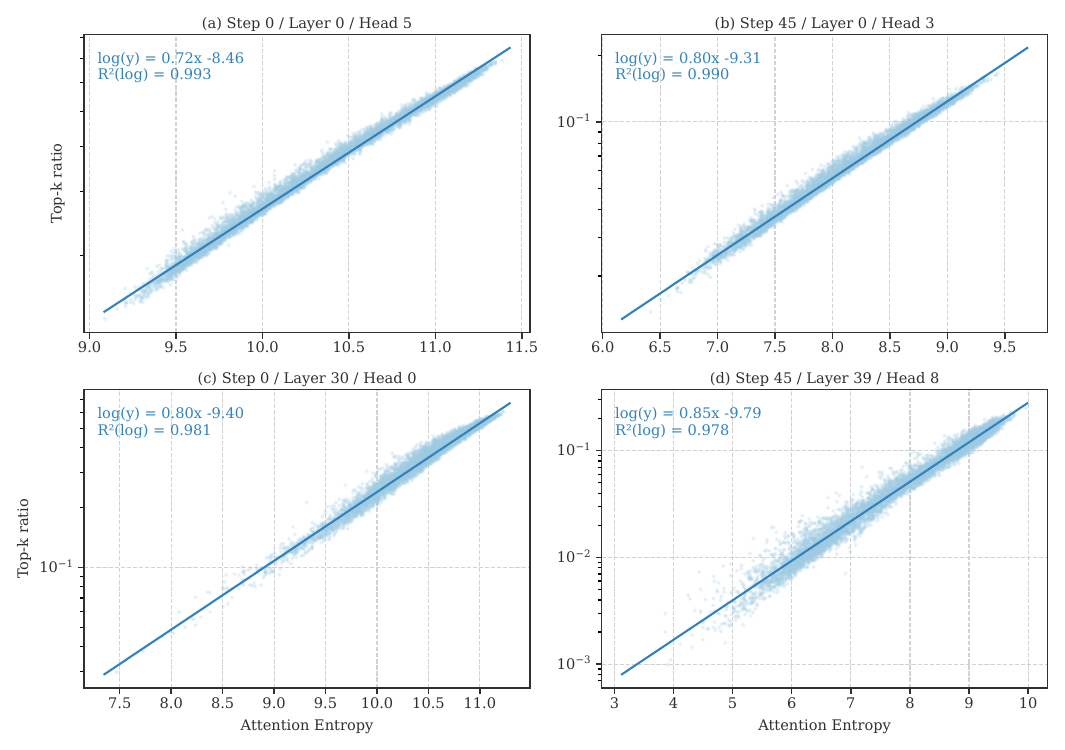}
    \end{minipage}
    \caption{
        Query-wise entropy--retained-density relation on HunyuanVideo-13B across
        different denoising steps, transformer layers, and attention heads.
        Each plot shows attention entropy against oracle retained density on
        a logarithmic vertical scale. Across all sampled configurations, the
        logarithm of retained density remains approximately linear in
        attention entropy, with \(R^2=0.978\)--\(0.993\). The consistently
        strong fits confirm that attention entropy provides a stable
        query-specific budget signal across model depths, attention heads,
        and denoising stages.
    }
    \label{fig:hunyuan_entropy_budget}
\end{figure*}

\paragraph{Entropy--budget relation.}
For each query, we compute its attention entropy \(\mathcal H_i\) and oracle
retained density \(B_i^\tau\), defined as the minimum fraction of top-ranked
keys required to preserve the target attention mass \(\tau\).
Figure~\ref{fig:hunyuan_entropy_budget} presents
representative fitting results across different denoising steps, transformer
layers, and attention heads. Across all sampled configurations, the logarithm
of the retained density remains approximately linear in attention entropy,
with \(R^2\) values ranging from \(0.978\) to \(0.993\). The fitted slopes also
remain consistent across the denoising trajectory. These results demonstrate
that entropy provides a stable query-specific budget signal across both
network components and denoising stages on HunyuanVideo.

\begin{figure*}[!t]
    \centering
    \begin{minipage}[t]{0.44\textwidth}
        \centering
        \includegraphics[width=\linewidth]{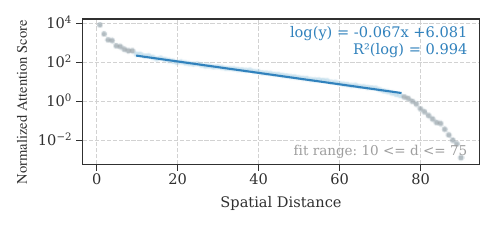}
        \vspace{1pt}

        \small\textbf{(a) Spatial-distance decay}
    \end{minipage}
    \hfill
    \begin{minipage}[t]{0.54\textwidth}
        \centering
        \includegraphics[width=\linewidth]
        {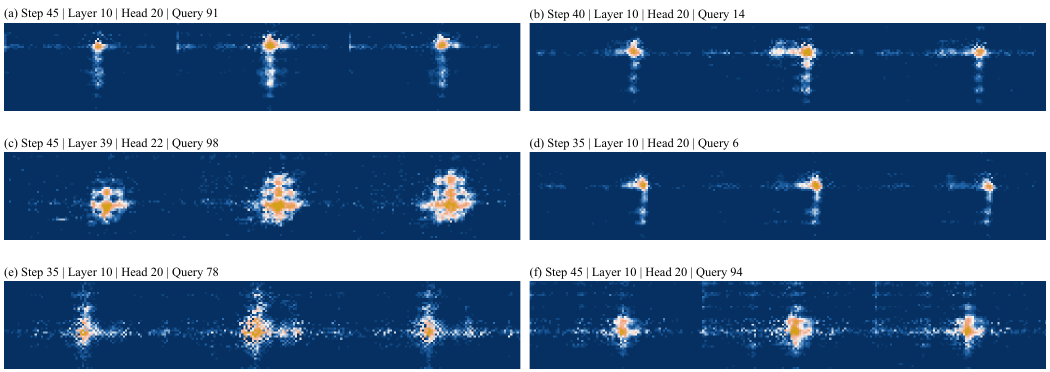}
        \vspace{1pt}

        \small\textbf{(b) Query-wise attention heatmaps}
    \end{minipage}
    \caption{
        Token Radius Attention pattern on HunyuanVideo-13B.
        \textbf{(a)} The normalized attention probability decays approximately
        exponentially with two-dimensional spatial distance, achieving
        \(R^2=0.994\) in log space over the fitted interval.
        \textbf{(b)} Representative query-wise heatmaps exhibit localized,
        query-centered supports with token-dependent spatial extents.
        These observations validate the query-centered spatial prior
        underlying Token Radius Attention beyond Wan-based models.
    }
    \label{fig:hunyuan_spatial_validation}
\end{figure*}

\paragraph{Token Radius Attention pattern.}
Figure~\ref{fig:hunyuan_spatial_validation}(a) aggregates normalized
attention probabilities according to the two-dimensional spatial distance between
query and key tokens. Within the fitted range
\(10\leq d_{\mathrm{s}}\leq75\), the logarithm of the normalized attention
probability follows
\(\log y=-0.067d_{\mathrm{s}}+6.081\), with \(R^2=0.994\).
The corresponding attention heatmaps in
Figure~\ref{fig:hunyuan_spatial_validation}(b) further show that dominant
interactions form query-centered spatial neighborhoods whose extent varies
across queries. Together, these results show that both token-specific
attention sparsity and the Token Radius Attention pattern generalize beyond
Wan-based models.

\subsection{Distributional Analysis of the Entropy-to-Budget Mapping}
\label{app:distributional_validation}

\paragraph{Effective-support approximation.}
The entropy-to-budget mapping used in the main paper is derived from an
\emph{entropy-equivalent uniform-support approximation}. Importantly, this
approximation does not assume that the original attention probabilities are
uniformly distributed. Instead, it replaces each query distribution with a
surrogate that uniformly distributes its probability mass over
\(\exp(\mathcal{H})\) effective keys while preserving the same entropy.
For a target attention mass \(\tau\), this gives the parameter-free mapping
\begin{equation}
    \widehat{B}_{\mathrm{Uni}}^\tau(\mathcal{H})
    =
    \frac{\tau\exp(\mathcal{H})}{N}.
    \label{eq:appendix_uniform_mapping}
\end{equation}
Although idealized, this approximation provides a simple monotonic relation
between query entropy and computation demand. We next examine whether this
relation remains valid under more realistic parametric models of attention
distributions and quantify the approximation error introduced by the
uniform-support surrogate.

\paragraph{General distributional formulation.}
Let \(X_1,\ldots,X_N\) be positive unnormalized attention weights drawn from
a parametric distribution \(F_{\boldsymbol{\theta}}\), and define
\begin{equation}
    p_j
    =
    \frac{X_j}{\sum_{\ell=1}^{N}X_\ell}.
\end{equation}
Let
\(\mu_{\boldsymbol{\theta}}
=\mathbb{E}_{\boldsymbol{\theta}}[X]\).
For sufficiently large \(N\), the entropy of the normalized attention
distribution can be approximated by
\begin{equation}
    \mathcal{H}(\boldsymbol{\theta})
    =
    \log N
    +
    \log\mu_{\boldsymbol{\theta}}
    -
    \frac{
        \mathbb{E}_{\boldsymbol{\theta}}[X\log X]
    }{
        \mu_{\boldsymbol{\theta}}
    }.
    \label{eq:distributional_entropy}
\end{equation}

For a retained key fraction \(q\), let
\(x_q=F_{\boldsymbol{\theta}}^{-1}(1-q)\).
The expected attention mass retained by the largest \(qN\) weights is
\begin{equation}
    L_{\boldsymbol{\theta}}(q)
    =
    \frac{
        \mathbb{E}_{\boldsymbol{\theta}}
        \left[
            X\mathbf{1}\{X\geq x_q\}
        \right]
    }{
        \mu_{\boldsymbol{\theta}}
    }.
    \label{eq:distributional_tail_mass}
\end{equation}
The corresponding retained density required to preserve target attention
mass \(\tau\) is
\begin{equation}
    B_{\boldsymbol{\theta}}^\tau
    =
    \inf
    \left\{
        q:
        L_{\boldsymbol{\theta}}(q)\geq\tau
    \right\}.
    \label{eq:distributional_budget}
\end{equation}
Therefore, an entropy-to-budget mapping can be obtained by identifying the
distribution parameters from entropy and substituting them into
Equation~\eqref{eq:distributional_budget}.

\paragraph{Parametric models of attention distributions.}
We fit \(44{,}400\) query-wise attention distributions collected from
Wan2.2-14B T2V under the \(40\)-step denoising setting. The observed
attention distributions can be approximated by several positive parametric
families, including Lognormal, Gamma, Weibull, Burr XII, and Johnson SU.
However, the best-fitting family varies across queries and entropy ranges,
rather than remaining fixed globally.

For one-shape scale families such as Lognormal, Gamma, and Weibull,
normalization eliminates the scale parameter and leaves one effective shape
parameter. This parameter can be identified from entropy through either a
closed-form expression or one-dimensional numerical inversion, yielding an
entropy-only budget mapping. More flexible families such as Burr XII and
Johnson SU retain multiple scale-free shape parameters. For these families,
entropy provides one constraint, while the remaining degrees of freedom must
be estimated from offline distribution statistics.

\begin{figure*}[!t]
    \centering
    \includegraphics[width=0.95\textwidth]{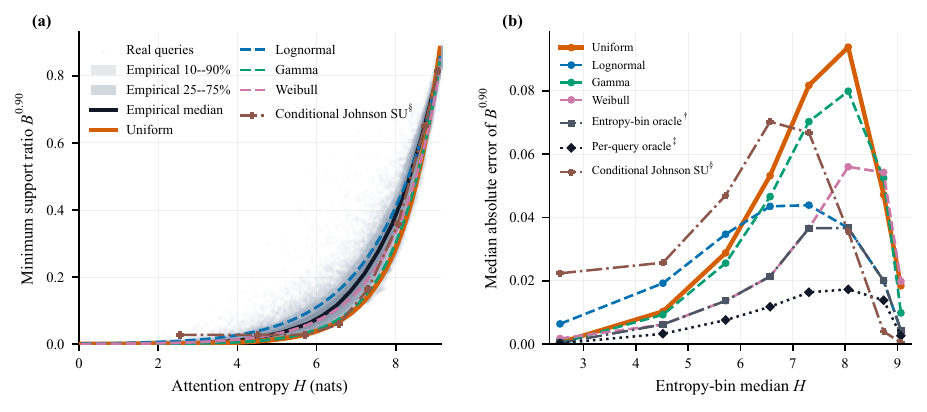}
    \caption{
        Distributional validation of the entropy-to-budget mapping at
        \(\tau=0.9\).
        \textbf{(a)} Oracle retained densities from real attention queries,
        empirical conditional intervals, and mappings derived under different
        distributional assumptions.
        Uniform, Lognormal, Gamma, and Weibull recover a consistent monotonic
        relation between entropy and the required retained density.
        Conditional Johnson SU uses offline statistics fitted on the complete
        query set.
        \textbf{(b)} Median absolute budget-prediction error within each
        entropy bin.
        The entropy-bin and per-query oracles select the best entropy-only
        mapping using ground-truth budgets and are included only as reference
        lower bounds.
    }
    \label{fig:distributional_budget_validation}
\end{figure*}

\begin{table*}[!t]
\centering
\small
\setlength{\tabcolsep}{5.0pt}
\renewcommand{\arraystretch}{1.05}
\begin{tabular*}{\textwidth}{@{\extracolsep{\fill}}llcccc@{}}
\toprule
\textbf{Mapping}
& \textbf{Parameter source}
& \textbf{MAE} $\downarrow$
& \textbf{Median AE} $\downarrow$
& \textbf{P90 AE} $\downarrow$
& \textbf{Mean error} \\
\midrule
Uniform
& Entropy only; closed form
& 0.0460 & 0.0282 & 0.1125 & -0.0415 \\
Lognormal
& Entropy only; analytic
& \textbf{0.0299} & 0.0205 & \textbf{0.0683} & +0.0164 \\
Gamma
& Entropy only; analytic
& 0.0420 & 0.0270 & 0.1024 & -0.0406 \\
Weibull
& Entropy only; analytic
& 0.0320 & \textbf{0.0196} & 0.0787 & -0.0243 \\
\midrule
Entropy-bin Oracle$^\dagger$
& Best mapping per entropy bin
& 0.0246 & 0.0122 & 0.0651 & -0.0064 \\
Per-query Oracle$^\ddagger$
& Best mapping per query
& 0.0127 & 0.0065 & 0.0296 & -0.0061 \\
Conditional Johnson SU$^\S$
& Entropy + offline statistics
& 0.0881 & 0.0277 & 0.2303 & +0.0261 \\
\bottomrule
\end{tabular*}
\caption{
Budget-prediction errors at $\tau=0.9$.
Mean error is defined as
$\mathbb{E}[\widehat B^{0.90}-B^{0.90}]$,
where positive values indicate over-estimation.
$^\dagger$ and $^\ddagger$ are oracle references among the four
entropy-only mappings and are not available during inference.
$^\S$ Conditional Johnson SU uses fitted offline statistics and is evaluated
on all 44,400 queries as a post-hoc reference.
}
\label{tab:distributional_budget_error}
\end{table*}

\paragraph{Entropy-identified Lognormal mapping.}
Consider
\(X\sim\operatorname{Lognormal}(\mu,\sigma^2)\).
Its relevant moments satisfy
\begin{equation}
    \mathbb{E}[X]
    =
    \exp\left(\mu+\frac{\sigma^2}{2}\right),
    \qquad
    \frac{\mathbb{E}[X\log X]}{\mathbb{E}[X]}
    =
    \mu+\sigma^2.
\end{equation}
Substituting them into
Equation~\eqref{eq:distributional_entropy} gives
\begin{equation}
    \mathcal{H}
    =
    \log N-\frac{\sigma^2}{2}.
\end{equation}
The scale parameter \(\mu\) disappears after normalization, and the
remaining parameter is uniquely determined by entropy:
\begin{equation}
    \sigma(\mathcal{H})
    =
    \sqrt{2(\log N-\mathcal{H})}.
    \label{eq:lognormal_sigma_appendix}
\end{equation}

Let \(\bar{\Phi}(z)=1-\Phi(z)\). The expected mass retained by the largest
fraction \(q\) of Lognormal samples is
\begin{equation}
    L_{\mathrm{LN}}(q)
    =
    \bar{\Phi}
    \left(
        \Phi^{-1}(1-q)-\sigma
    \right).
\end{equation}
Solving \(L_{\mathrm{LN}}(q)=\tau\) yields
\begin{equation}
    B_{\mathrm{LN}}^\tau(\mathcal{H})
    =
    \bar{\Phi}
    \left(
        \sqrt{2(\log N-\mathcal{H})}
        -
        \Phi^{-1}(\tau)
    \right).
    \label{eq:lognormal_budget_appendix}
\end{equation}
Thus, the Lognormal assumption also produces an analytic entropy-to-budget
mapping without requiring a learned budget predictor. Gamma and Weibull
similarly yield entropy-only mappings through one-dimensional parameter
inversion.

\paragraph{Empirical comparison.}
Figure~\ref{fig:distributional_budget_validation}(a) shows that oracle
retained density increases monotonically with attention entropy
(Spearman \(\rho=0.9884\)). Uniform-support, Lognormal, Gamma, and Weibull
models recover the same dominant trend despite different tail assumptions,
confirming that entropy is the primary signal for query-specific attention
demand. Table~\ref{tab:distributional_budget_error} shows that the
uniform-support approximation has an MAE of \(0.0460\), only \(0.0161\)
higher than the best individual entropy-only mapping. More flexible oracle
or conditional models either require ground-truth budgets or offline fitted
statistics and are not available in TRA's inference pipeline. We therefore
retain the uniform-support mapping because it is closed form, parameter free,
and consistent with the empirical entropy--budget relation.

\subsection{Cross-Step Stability of Attention Entropy}
\label{app:entropy_stability}

\paragraph{Entropy stability across denoising steps.}
TRA estimates query-wise attention entropy during the dense warm-up stage
and reuses the resulting token budgets and radius maps in subsequent sparse
denoising steps, following prior observations of cross-step entropy
stability~\cite{chen2026ecovideo}. To validate this design, for each
transformer block and attention head, we flatten the entropy values of all
queries at each denoising step into an entropy vector and compute pairwise
cosine similarities between different steps.

\begin{figure*}[!t]
    \centering
    \includegraphics[width=0.79\textwidth]{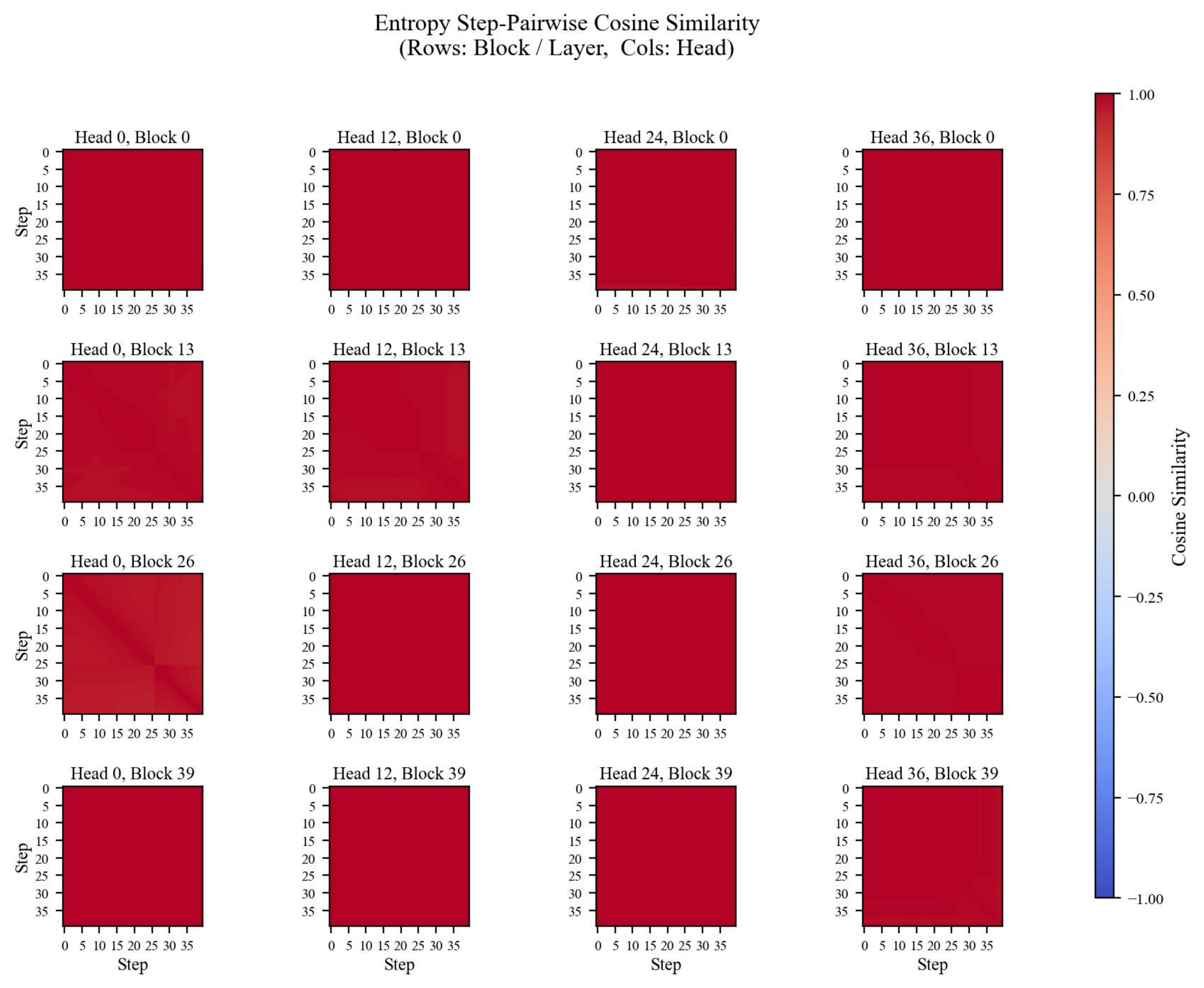}
    \caption{
        Cross-step stability of query-wise attention entropy.
        Each matrix reports the pairwise cosine similarity between entropy
        vectors at different denoising steps for a representative transformer
        block and attention head. The similarities remain close to \(1\),
        with a minimum value of \(0.945\).
    }
    \label{fig:entropy_step_similarity}
\end{figure*}

As shown in Figure~\ref{fig:entropy_step_similarity}, the query-wise entropy
patterns remain highly consistent across denoising steps for representative
network depths and attention heads. The cosine similarities are consistently
close to \(1\), and even the least similar step pair achieves a similarity of
\(0.945\). These results indicate that, although the absolute entropy values
may evolve during denoising, their relative distribution across queries
remains largely stable. Since TRA adopts a monotonic entropy-to-budget
mapping, this stability supports reusing the token budgets and attention
radii estimated during dense warm-up.

\section{Supplementary Experiments}
\label{app:supplementary_experiments}

\subsection{Hyperparameter Sensitivity}
\label{app:hyperparameter_ablation}

\paragraph{Experimental settings.}
We conduct all hyperparameter ablations on Wan2.2-14B for T2V generation
with \(40\) denoising steps at \(720\times1280\) resolution. Unless otherwise
specified, TRA uses \(10\) dense warm-up steps, one prefix dense transformer
layer, a block size of \(90\), and no periodic entropy refresh. PSNR, SSIM,
and LPIPS are measured against the corresponding Dense Attention outputs.

\paragraph{Hyperparameter sensitivity.}
As shown in Table~\ref{tab:hyperparameter_ablation}, increasing the dense
warm-up length consistently improves reconstruction fidelity at the cost of
moderately increased runtime. In particular, increasing the warm-up length
from \(10\) to \(20\) steps improves PSNR from \(23.452\) to \(29.214\),
while increasing runtime from \(1015\) s to \(1234\) s. Increasing the number
of dense layers or distributing them uniformly across the network provides
no consistent quality improvement, supporting the use of a single prefix
dense layer.

Increasing the block size generally reduces runtime, whereas an excessively
large block size of \(120\) noticeably degrades reconstruction fidelity.
A block size of \(90\) therefore provides a favorable quality--efficiency
trade-off. Together with the cross-step stability in
Section~\ref{app:entropy_stability}, these results support reusing the entropy
estimated during dense warm-up without periodic refresh.

\begin{table*}[!t]
  \centering
  \begingroup
  \fontsize{7.8}{8.9}\selectfont
  \renewcommand{\arraystretch}{0.96}

  \begin{minipage}[t]{0.485\textwidth}
    \centering
    \setlength{\tabcolsep}{2.7pt}
    \begin{tabular*}{\linewidth}{@{\extracolsep{\fill}}lrrrr@{}}
      \toprule
      \textbf{Setting}
      & \textbf{Time}
      & \textbf{PSNR}
      & \textbf{SSIM}
      & \textbf{LPIPS} \\
      \midrule
      \multicolumn{5}{l}{\textit{Dense Warm-up Steps}} \\
      \textbf{10} & 1015 & 23.4520 & 0.8020 & 0.1615 \\
      11 & 1043 & 24.2380 & 0.8163 & 0.1504 \\
      12 & 1075 & 24.8880 & 0.8287 & 0.1346 \\
      13 & 1104 & 25.7220 & 0.8411 & 0.1214 \\
      14 & 1105 & 26.5170 & 0.8562 & 0.1053 \\
      15 & 1128 & 27.2420 & 0.8674 & 0.0958 \\
      16 & 1146 & 27.5740 & 0.8714 & 0.0927 \\
      17 & 1167 & 27.7680 & 0.8749 & 0.0896 \\
      18 & 1195 & 28.3360 & 0.8831 & 0.0814 \\
      19 & 1228 & 28.8090 & 0.8895 & 0.0763 \\
      20 & 1234 & 29.2140 & 0.8946 & 0.0716 \\
      \midrule
      \multicolumn{5}{l}{\textit{Entropy and Mask Refresh}} \\
      \textbf{Never} & 1018 & 21.6030 & 0.7589 & 0.2035 \\
      Every 10 Steps & 1112 & 21.6030 & 0.7589 & 0.2035 \\
      Every 5 Steps & 1198 & 21.6030 & 0.7589 & 0.2035 \\
      Every 2 Steps & 1481 & 21.6030 & 0.7589 & 0.2035 \\
      \bottomrule
    \end{tabular*}
  \end{minipage}
  \hfill
  \begin{minipage}[t]{0.485\textwidth}
    \centering
    \setlength{\tabcolsep}{2.7pt}
    \begin{tabular*}{\linewidth}{@{\extracolsep{\fill}}lrrrr@{}}
      \toprule
      \textbf{Setting}
      & \textbf{Time}
      & \textbf{PSNR}
      & \textbf{SSIM}
      & \textbf{LPIPS} \\
      \midrule
      \multicolumn{5}{l}{\textit{Number and Placement of Dense Layers}} \\
      \textbf{1, Prefix} & 1036 & 23.4520 & 0.8020 & 0.1615 \\
      1, Uniform & 1016 & 23.4520 & 0.8020 & 0.1615 \\
      2, Prefix & 1030 & 23.5820 & 0.8053 & 0.1490 \\
      2, Uniform & 1036 & 23.4690 & 0.8001 & 0.1622 \\
      3, Prefix & 1067 & 23.2840 & 0.8044 & 0.1488 \\
      3, Uniform & 1043 & 23.3890 & 0.7962 & 0.1658 \\
      4, Prefix & 1057 & 23.2200 & 0.8033 & 0.1534 \\
      4, Uniform & 1060 & 23.4270 & 0.7993 & 0.1626 \\
      \midrule
      \multicolumn{5}{l}{\textit{Block Size}} \\
      40 & 1295 & 23.3930 & 0.7957 & 0.1690 \\
      48 & 1211 & 23.3710 & 0.7923 & 0.1734 \\
      72 & 1068 & 23.6530 & 0.7957 & 0.1686 \\
      \textbf{90} & 1019 & 23.4520 & 0.8020 & 0.1615 \\
      120 & 952 & 22.2880 & 0.7742 & 0.1767 \\
      \bottomrule
    \end{tabular*}
  \end{minipage}

  \endgroup
  \caption{
      Additional hyperparameter ablations on Wan2.2-14B T2V with
      \(40\) denoising steps at \(720\times1280\) resolution.
      Time is reported in seconds. Unless otherwise specified, TRA uses
      \(10\) dense warm-up steps, one prefix dense layer, block size \(90\),
      and no periodic entropy refresh. Reconstruction metrics are measured
      against Dense Attention outputs.
  }
  \label{tab:hyperparameter_ablation}
\end{table*}

\subsection{Additional Qualitative Results}
\label{app:additional_qualitative}

\begin{figure*}[!t]
    \centering
    \includegraphics[width=\textwidth]{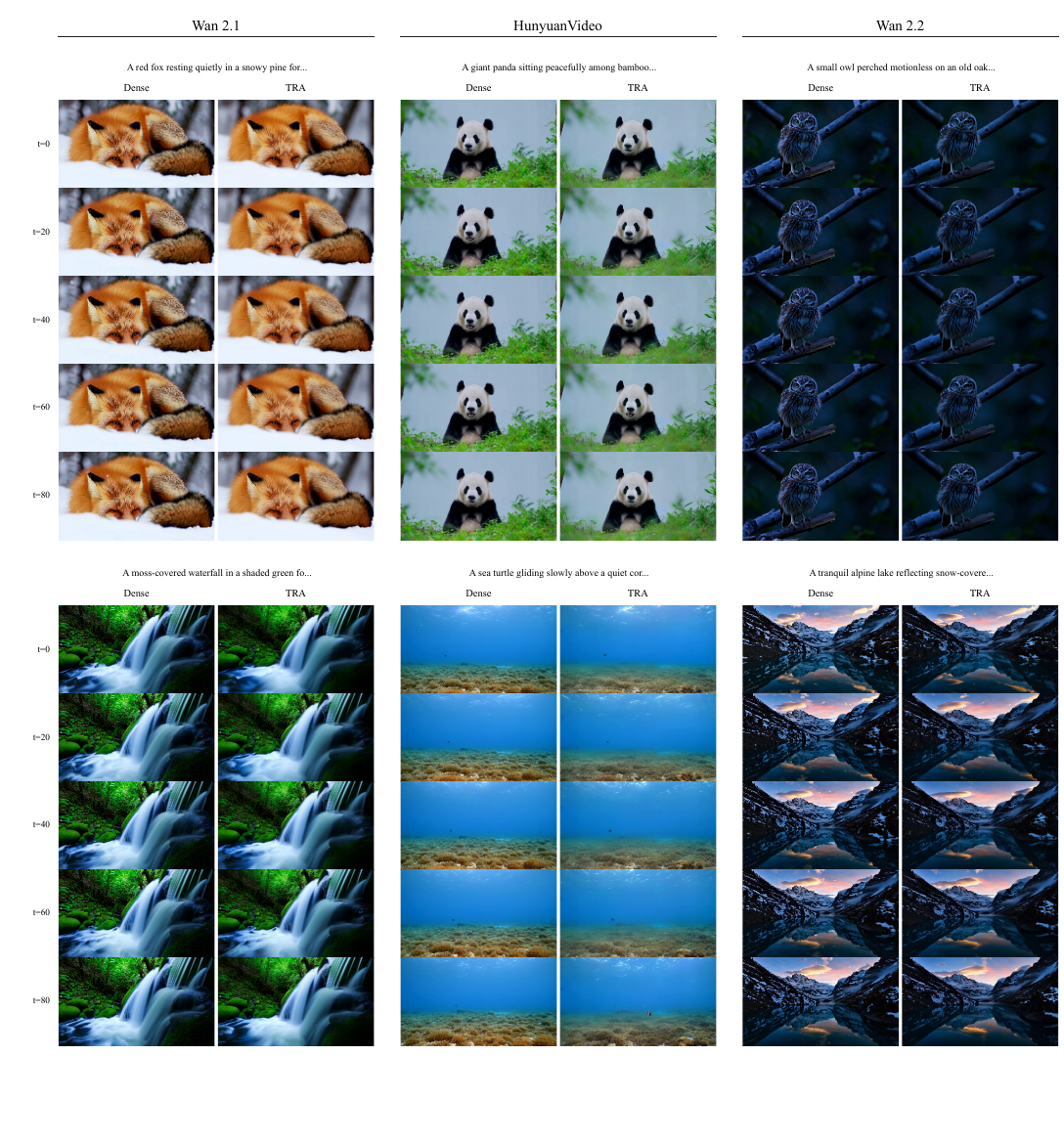}
    \caption{
        Additional qualitative results generated with TRA.
        Each row shows temporally ordered frames sampled from one video.
        TRA preserves coherent appearance, scene structure, and motion
        throughout the generated sequence.
    }
    \label{fig:additional_qualitative_results}
\end{figure*}

Figure~\ref{fig:additional_qualitative_results} presents additional
frame-level visualizations generated with TRA. Each row contains temporally
ordered frames sampled from one video. TRA preserves subject appearance,
scene structure, fine-grained details, and temporal consistency while using
query-specific sparse attention.

\section{Implementation Details}
\label{app:kernels}

Let \(F\) be the number of latent video frames,
\(P=HW\) the number of spatial tokens per frame,
and \(N=FP\) the number of video tokens.
Let \(S\) denote the total sequence length, including optional text tokens.
The sequence is partitioned into
\[
    B=\left\lceil\frac{S}{b}\right\rceil
\]
blocks of \(b\) tokens. The mask-construction kernel outputs a Boolean
block mask, where an entry of one denotes a retained query--key block pair.
We use \(H_a\) to denote the number of attention heads.

\subsection{Flash Radius Mask Construction}
\label{app:mask_kernel}

\paragraph{Entropy-to-budget-to-radius lookup.}
For attention head \(h\) and query token \(q\), the fused entropy kernel
produces the query-wise attention entropy \(\mathcal H_{h,q}\). Following the
main method, TRA first estimates the target key count and then obtains the
smallest boundary-aware base radius whose discrete support reaches that count:
\begin{equation}
    \widehat K_{h,q}^{\tau}
    =
    \left\lceil\tau\exp(\mathcal H_{h,q})\right\rceil,
    \qquad
    r_{h,q}
    =
    \mathcal L
    \left(f_q,x_q,y_q,\widehat K_{h,q}^{\tau}\right),
    \label{eq:implementation-budget-radius}
\end{equation}
where \(\tau\) is the target attention mass and \(\mathcal L\) implements the
boundary-aware radius selection defined by the discrete support count in the
main method.
For fixed latent-grid dimensions and temporal decay, the discrete counts are
precomputed; inference therefore requires only the analytic budget evaluation
and a lookup, without per-query key ranking or iterative radius search.

\paragraph{Temporal radius decay.}
The implementation uses the same exponential temporal decay as the main
method~\cite{li2026radial},
\(\phi(\delta)=\exp(-\gamma\delta)\). Thus, for a query--key pair \((q,k)\),
the spatial threshold is
\begin{equation}
    \rho_{h,q,k}
    =
    r_{h,q}\phi(|f_q-f_k|).
    \label{eq:kernel-radius-threshold}
\end{equation}

\paragraph{Token-pair distance.}
Let \(\mathbf x_q=(x_q,y_q)\) denote the two-dimensional spatial coordinate
of token \(q\). Consistent with Insight~II and the mask definition in the
main paper, the kernel evaluates
\begin{equation}
    d_s(q,k)
    =
    \left\|
        \mathbf x_q-\mathbf x_k
    \right\|_2.
    \label{eq:implementation-spatial-distance}
\end{equation}
For the sequence-distance ablation, the spatial term is replaced by
\(|q-k|\). To match the area of a two-dimensional disk using a symmetric
one-dimensional interval, the corresponding radius is initialized as
\begin{equation}
    \rho_{\mathrm{1D}}
    =
    \frac{\pi}{2}
    \rho_{\mathrm{2D}}^2.
    \label{eq:area-matched-1d}
\end{equation}

\begin{algorithm}[t]
\caption{CUDA block-mask construction.}
\label{alg:mask-kernel}
\begin{algorithmic}[1]
\Require
attention entropy \(\mathcal H\in\mathbb R^{H_a\times N}\);
target attention mass \(\tau\);
boundary-aware radius lookup \(\mathcal L\);
block size \(b\);
number of video blocks
\(B_{\mathrm v}=\lceil N/b\rceil\)
\Ensure
block mask \(M\in\{0,1\}^{H_a\times B\times B}\)

\State
\(\widehat K\gets\lceil\tau\exp(\mathcal H)\rceil\),
\(R\gets\mathcal L(\widehat K)\)
\Comment{query-specific budgets and radii}

\ForAll{attention heads \(h\), query blocks \(u\), and key blocks \(v\)
        \textbf{in parallel}}

    \If{\(u\geq B_{\mathrm v}\) \textbf{or} \(v\geq B_{\mathrm v}\)}
        \State \(M_{h,u,v}\gets1\)
        \State \textbf{continue}
        \Comment{text or non-video blocks remain dense}
    \EndIf

    \If{\(\textsc{BoundingBoxReject}(h,u,v,R)\)}
        \State \(M_{h,u,v}\gets0\)
        \State \textbf{continue}
        \Comment{no token pair can satisfy the radius constraint}
    \EndIf

    \ForAll{key columns \(j=0,\ldots,b-1\) \textbf{in parallel}}
        \State \(c_j\gets0\)

        \ForAll{query lanes \(i=0,\ldots,b-1\) \textbf{in parallel}}
            \State \(q\gets ub+i\), \(k\gets vb+j\)

            \If{\(q<N\) \textbf{and} \(k<N\)}
                \State
                \(\rho\gets R_{h,q}\phi(|f_q-f_k|)\)

                \State
                \(z_{i,j}\gets[d_s(q,k)\leq\rho]\)

                \State
                \(c_j\gets c_j+z_{i,j}\)
                \Comment{warp ballot and population count}
            \EndIf
        \EndFor
    \EndFor

    \State
    \(\mathcal C\gets\{j:c_j>0\}\)
    \Comment{nonempty key columns}

    \State
    \(\mathcal C_{\mathrm{hi}}
    \gets\{j:c_j>b/3\}\)
    \Comment{high-coverage columns}

    \State
    \(M_{h,u,v}
    \gets
    [|\mathcal C|>0]
    \wedge
    [10|\mathcal C_{\mathrm{hi}}|>6|\mathcal C|]\)

    \Comment{retain if \(>60\%\) of nonempty columns have high coverage}
\EndFor
\end{algorithmic}
\end{algorithm}

\paragraph{Kernel execution.}
Before mask construction, each
\(H\times W\) frame is partitioned into spatial
tiles, and its video tokens are rearranged in tile-major order. Spatially
neighboring tokens therefore become contiguous in the one-dimensional
sequence, allowing token-radius supports to be represented by fewer and
denser query--key blocks. The same permutation is applied to \(Q\), \(K\),
and \(V\), and the attention outputs are restored to the original token order
after sparse attention.

The CUDA grid assigns one thread block to each \((h,u,v)\) tuple.
A bounding-box test first rejects block pairs that cannot contain any valid
token interactions. For the remaining pairs, one thread is assigned to each
query lane, while warp ballots and population counts compute the number of
valid query tokens for every key column. A single final write produces
\(M_{h,u,v}\).

The resulting block mask remains head and query specific. For joint
text--video attention, blocks containing text keys remain dense, matching the
text-key handling defined in the main method.

\subsection{Fused Attention Entropy Kernel}
\label{app:entropy_kernel}

For query row \(i\), let
\begin{equation}
    s_{ij}
    =
    \frac{\mathbf q_i^\top\mathbf k_j}{\sqrt d},
    \qquad
    p_{ij}
    =
    \frac{\exp(s_{ij})}
    {\sum_{\ell}\exp(s_{i\ell})}.
\end{equation}
The Shannon entropy in nats can be rewritten as
\begin{equation}
    \mathcal H_i
    =
    -\sum_jp_{ij}\log p_{ij}
    =
    \log
    \left(
        \sum_j e^{s_{ij}}
    \right)
    -
    \sum_jp_{ij}s_{ij}.
    \label{eq:fused-entropy}
\end{equation}
This identity allows entropy to be accumulated together with the online
softmax state without materializing the complete attention matrix.

\begin{algorithm}[t]
\caption{Fused online softmax and entropy extraction.}
\label{alg:entropy-kernel}
\begin{algorithmic}[1]
\Require
Query tile \(Q\);
key/value tiles
\(\{(K^{(r)},V^{(r)})\}_{r=1}^{R}\);
softmax scale \(\alpha=1/\sqrt d\)
\Ensure
Attention output \(O\);
per-query attention entropy \(\mathcal H\)

\ForAll{query rows \(i\) in the tile \textbf{in parallel}}
    \State
    \(m\gets-\infty\),
    \(\ell\gets0\),
    \(a\gets0\),
    \(o\gets\mathbf0\)

    \For{\(r=1\) to \(R\)}
        \State
        \(\mathbf s
        \gets
        \alpha\mathbf q_i(K^{(r)})^\top\)

        \State
        \(m_r\gets\max_j s_j\)

        \State
        \(\mathbf w\gets
        \exp(\mathbf s-m_r)\)

        \State
        \(\ell_r\gets\sum_jw_j\)

        \State
        \(a_r\gets\sum_jw_js_j\)

        \State
        \(\mathbf o_r\gets
        \sum_jw_j\mathbf v_j\)

        \State
        \(m'\gets\max(m,m_r)\)

        \State
        \(u\gets\exp(m-m')\),
        \(v\gets\exp(m_r-m')\)

        \State
        \(\ell\gets u\ell+v\ell_r\)

        \State
        \(a\gets ua+va_r\)

        \State
        \(o\gets uo+v\mathbf o_r\)

        \State
        \(m\gets m'\)
    \EndFor

    \State
    \(O_i\gets o/\ell\)

    \State
    \(\mathcal H_i\gets\log\ell+m-a/\ell\)

    \Comment{equivalent to Equation~\eqref{eq:fused-entropy}}
\EndFor
\end{algorithmic}
\end{algorithm}

\paragraph{Computational cost.}
The entropy accumulator reuses the same score tiles and online-softmax states
as the dense attention forward pass. It introduces only scalar per-row
accumulators and one entropy writeback, rather than a second attention pass
or explicit softmax materialization. Consequently, entropy extraction does
not change the asymptotic complexity of dense attention and requires only
\(O(H_aN)\) additional storage.

Let \(B_{\mathrm v}\) be the number of video-token blocks. The worst-case
logical complexity of mask construction is
\begin{equation}
    O\!\left(
        H_aB_{\mathrm v}^{2}b^{2}
    \right).
\end{equation}
In practice, the bounding-box test rejects spatially incompatible block
pairs before token-level evaluation, while tile-major ordering concentrates
valid interactions into a small number of neighboring blocks. The final mask
requires \(O(H_aB^2)\) Boolean storage and is consumed directly by the
FlashInfer block-sparse attention backend~\cite{ye2025flashinfer}.
\section{Limitations and Future Work}
\label{app:limitations}

\paragraph{Restricted and static sparsity patterns.}
TRA uses a spatiotemporal distance prior to convert query-specific token
budgets into structured sparse supports. Although the radius is adapted to
the entropy of each query, the resulting support is still constrained to a
predefined, distance-based pattern with a fixed temporal-decay rule. Such a
pattern may not fully capture content-dependent interactions, including
long-range object correspondence, nonlocal motion, or scene-dependent
attention structures. Future work could explore online token selection or
richer adaptive patterns that jointly account for semantic content, motion,
and spatial structure.

\paragraph{Lack of a dedicated token-level sparse-attention kernel.}
TRA currently executes query-specific token-level sparsity through an
existing block-sparse attention backend. Tile-major token reordering and
block-mask construction reduce blockification overhead but retain redundant
computation within selected blocks. A dedicated kernel that fuses mask
construction, token gathering, and sparse attention could further improve
the practical efficiency of TRA.

\end{document}